\documentclass[10pt,sigconf]{acmart}
\usepackage{grffile}
\usepackage{algorithm}
\usepackage[algo2e, ruled, vlined]{algorithm2e}
\newcommand{\ignore}[1]{}
\newcommand{\notinproc}[1]{#1}
\newcommand{\onlyinproc}[1]{}
\newcommand{\Walks}{\mathop{\texttt{{\sf Walk}}}}
\newcommand{\Pairs}{\mathop{\texttt{{\sf Pairs}}}}
\newcommand{\Lscore}{\mathop{\texttt{{\sf Lscore}}}}
\newcommand{\E}{\mathop{\texttt{{\sf E}}}}
\newcommand{\nepochs}{\mathop{\texttt{{\sf \#epochs}}}}
\AtBeginDocument{%
  \providecommand\BibTeX{{%
    \normalfont B\kern-0.5em{\scshape i\kern-0.25em b}\kern-0.8em\TeX}}}

\notinproc{
\setcopyright{none}
}
\onlyinproc{
\setcopyright{rightsretained}
}

\usepackage{xcolor} 

\SetCommentSty{mycommfont}

\begin{document}
\copyrightyear{2020}
\acmYear{2020}
\acmConference[GRADES-NDA'20]{3rd Joint International Workshop on Graph Data Management Experiences \& Systems (GRADES) and Network Data Analytics (NDA)}{June 14, 2020}{Portland, OR, USA}
\acmBooktitle{3rd Joint International Workshop on Graph Data Management Experiences \& Systems (GRADES) and Network Data Analytics (NDA) (GRADES-NDA'20), June 14, 2020, Portland, OR, USA} \acmDOI{10.1145/3398682.3400060} \acmISBN{978-1-4503-8021-8/20/06}

\title{Graph Learning with Loss-Guided Training}

\author{Eliav Buchnik}
\email{eliavbuh@gmail.com}
\authornotemark[1]
\affiliation{%
\institution{Tel Aviv University}
\institution{Google Research}
}

\author{Edith Cohen}
\email{edith@cohenwang.com}
\authornotemark[1]
\affiliation{%
\institution{Google Research}
\institution{Tel Aviv University}
}

\begin{abstract}
Classically, ML models trained with stochastic gradient descent (SGD) are designed to minimize the average loss per example and use a distribution of training examples that remains {\em static} in the course of training.  Research in recent years demonstrated, empirically and theoretically, that significant acceleration is possible by methods that dynamically adjust the training distribution in the course of training so that training is more focused on examples with higher loss.
We explore {\em loss-guided training} in a new domain of node embedding methods pioneered by {\sc DeepWalk}.  These methods work with implicit and large set of positive training examples that are generated using random walks on the input graph and therefore are not amenable for typical example selection methods.  We propose computationally efficient methods that allow for loss-guided training in this framework. Our empirical evaluation on a rich collection of datasets shows significant acceleration over the baseline static methods, both in terms of total training performed and overall computation.
\end{abstract}

\onlyinproc{

\ccsdesc[500]{Computing methodologies~Dimensionality reduction and manifold learning}
}

\maketitle

\section{INTRODUCTION}
Graph data is prevalent and models entities (nodes) and interactions between them (edges). The edges may corresponds to provided interactions between entities (likes, purchases, messages, hyperlinks) or are derived from metric data, for example,  by connecting each point to its nearest neighbors.

{\em Node embeddings}, which are representations of graph nodes in the form of low dimensional vectors, are an important component in graph analysis pipelines. They are used as task-agnostic representation with downstream tasks that include node classification,
node clustering for community detection, and link prediction for recommendations \cite{deepwalk:KDD2014, node2vec:kdd2016, DBLP:journals/tkde/CaiZC18}.
Embeddings are computed with the qualitative objective of preserving  structure -- so that nodes that are more connected get assigned closer embedding vectors~\cite{imagenet:CVPR2009, Koren:kdd2008, Mikolov:NIPS13, deepwalk:KDD2014, BERRY95, PinSage:KDD2018}.
The optimization objective  has the general form of a weighted sum over examples (pairs of nodes) of a per-example loss function and are commonly performed using  stochastic gradient descent (SGD) \cite{SGDbook:1971,Koren:kdd2008,Salakhutdinov:ICML2007,Gemulla:KDD2011,Mikolov:NIPS13}.
  
\subsection{Node embeddings via random walks}
A first attempt to obtain positive training examples (node pairs) from the input graph is to use the provided set of edges \cite{Koren:IEEE2009}.  
A highly effective approach, pioneered by {\sc DeepWalk} \cite{deepwalk:KDD2014}, is to instead select examples based on co-occurrence of pairs in short random walks performed on the input graph. These methods weight and greatly expand the set of positive examples.
{\sc DeepWalk} treats random walks on the graph as "sentences" of nodes and applies the popular word embedding framework {\sc Word2Vec} \cite{Mikolov:NIPS13}.
{\sc Node2vec} \cite{node2vec:kdd2016} further refined the method by extending the family of random walks with hyper-parameters that tune the depth and breadth of the walk. Prolific followup work (see summary in \cite{DBLP:journals/tkde/CaiZC18, Wu_2020}) further extended the family of random walks but retained the general structure of producing "node sentences." 

\subsection{Loss-guided training}
Random-walk base methods were studied in settings where the distribution of random walks and thus the distribution of training examples remain {\em static} in the course of training.
A prolific research thread proposed methods that accelerate the training or improve accuracy by {\em dynamically} modifying the distribution of examples in the course of training~\cite{curriculumlearning:ICML2009,AlainBengioSGD:2015,ZhaoZhang:ICML2015,Shrivastava:CVPR2016,facenet:cvpr2015,LoshchilovH:ICLR2016,shalev:ICML2016}:
 These approaches include 
Curriculum/self-paced learning \cite{curriculumlearning:ICML2009}, where the selection is altered to mimic human learning: First the algorithm learns over the "easy" examples and then moves to "hard" examples, where margin is used as a measure of difficulty. A related approach guides the example
selection process by the current magnitude of the gradient or the loss value.  One proposed method applies importance sampling according to loss or gradient
\cite{AlainBengioSGD:2015,ZhaoZhang:ICML2015}, which preserves the expected value of the stochastic gradient updates but spreads them differently.
Other methods focus on higher loss examples in a biased fashion
that essentially alters the objective: Hard examples for image training \cite{Shrivastava:CVPR2016,facenet:cvpr2015}, selecting examples by moving average of the loss \cite{LoshchilovH:ICLR2016}, or focusing entirely on the highest loss examples \cite{shalev:ICML2016} with a compelling theoretical underpinning,
 Overall, these methods were studied with supervised learning and as far as we know, were not explored for computing node embeddings.

\subsection{Our contribution}
We propose and study methods that incorporate dynamic training, in particular example selection that is focused on higher loss examples,
in the particular context of popular random-walk based example selection methods for node embedding.  The hope is that we can obtain similar gains in performance as observed in other domains. 

 The application of loss-guided training to random-walk based methods poses some methodical and computational challenges. 
 First, the methods used for other domains are not directly applicable.  They were considered in supervised situations where the input data has the form of (example, label) pairs which are available explicitly and make the loss computation straightforward.  In our setting,
examples are produced during training using random walks: The potential number of examples can be quadratic in the number of nodes even when the input graph is sparse and the set is implicit in the graph representation. Thus,  per-example state or loss evaluation on all potential examples cannot be efficiently maintained, which rules out  approaches such as 
\cite{shalev:ICML2016,LoshchilovH:ICLR2016}.

 Second, dynamic example selection, and in particular loss-guided example selection, tends to be computation heavy and trades-off the efficiency of {em training} (performing the gradient updates) and efficiency of {\em preprocessing} (the computation needed to generate the training  sequence~\cite{AlainBengioSGD:2015,ZhaoZhang:ICML2015,LoshchilovH:ICLR2016}).
Even with the baseline random walk methods, the computational resources needed increase with graph size, the length and type of the random walk, the number of generated examples from the walk, and the dimension of the embedding vectors. In practice, the cost of embedding computation tends to be a significant part of the overall downstream pipeline.
We aim to enhance random-walk based methods without compromising their scalability.  
 
The components of training and preprocessing costs  typically draw on different resources (e.g., gradient updates are communicated). We aim for efficiency and design loss-guided training methods that provide tunable trade-offs.  Our most effective approaches work with the same random walk processes as the respective baseline methods and assign {\em loss scores} to walks (each generating a set of examples) instead of to individual examples. At each selection phase we generate a set of random walks according to the baseline model, assign loss scores to these walks (via methods to be detailed later on), and choose a sample of the walks for training that is weighted by their loss scores.  We empirically show that across a variety of datasets, our loss-guided methods provide dramatic reduction in training cost with a very small increase in preprocessing cost compared with the baseline methods that use a static distribution of training examples.

\subsection{Related work}
Graph Neural Networks (GNNs) are an emerging approach
for graph learning tasks (see survey \cite{GNNsurvey:2020}). Notably, 
Graph Convolutional Networks \cite{AtwoodT:NIPS2016,DefferrardBV:NIPS2016,KipfW:ICLR2017} work with node features and create representations in terms of node features \cite{PinSage:KDD2018}.
Variational auto-encoders \cite{kipf2016variational} produce node embeddings in an unsupervised fashion but perform similarly to prior methods.
Random-walk based methods remain a viable alternative that obtains state of the art results for node representations computed from graph structure alone.

\subsection{Overview}
The paper is organized as follows. In Section~\ref{prelim:sec} we provide necessary background on the baseline node embedding methods
{\sc DeepWalk} \cite{deepwalk:KDD2014} and {\sc Node2Vec} \cite{node2vec:kdd2016} and the Word2Vec SGNS framework \cite{Mikolov:NIPS13} that they build on.  In Section 
\ref{lossguidedalgs:sec} we present our methods that dynamically modify the distribution of training examples according to loss.
We provide details on our experimental setup in Section~\ref{sec:empirical_evaluation}. We illustrate the benefits of loss-guided training using a synthetic example network in
Appendix \ref{case_study:sec}. The real-life datasets and tasks used in our experiments are described in Section \ref{datasets:sec} and results are reported and discussed in Section \ref{results:sec} and 
 Appendix~\ref{param_sensitivity:sec}-\ref{loss:sec}.

\section{Preliminaries} \label{prelim:sec}
We consider graph datasets of the form $(V,E,w)$ with a set of nodes $V$, that represent entities, a set of edges $E\subset V\times V$ that represent pairwise interactions, and an assignment $w$ of positive scalar weights to edges that correspond to the strength of interactions.
\notinproc{
Entities may be of different types (for example, users and videos with edges corresponding to views) or be of the same type (words in a text corpus with edges corresponding to co-occurrences or users in a social network and edges corresponding to interactions).}
A node embeddings is a mapping of nodes $i\in V$ to vectors $\boldmath{f}_{i} \in \Re^{d}$, where typically $d << |V|$. 

\subsection{Overview of baseline methods}
The node embeddings methods we consider here are based on the popular {\sc DeepWalk} \cite{deepwalk:KDD2014} and  its refinement {\sc Node2vec} \cite{node2vec:kdd2016}. 
Algorithm~\ref{alg:baseline} provides a high-level view of the baseline methods.
These methods build on the {\sc word2vec} \cite{Mikolov:NIPS13} Skip Gram with Negative Sampling (SGNS) method.  SGNS was originally designed for learning embeddings for words in a text corpus. The method generates short sequences (referred to as {\em sentences} of consecutive words from the text corpus and uses these sentences for training (more details are provided below).
The node embeddings methods generate instead sequences of nodes using short random walks on the graph and apply the SGNS framework to these node "sentences" in a black box fashion. 
The node embedding methods differ in the distribution over node sentences.  Both our baselines specify distributions $\Walks[i,t]$ of random walks of length $t$ that start from a node $i$.  
{\sc DeepWalk} conducts a simple random walk, where the next node is selected independently of history according to weight of outgoing edges, that is, if the walk is node $i$ then the probability of continuing to node $j$ is $w_{ij}/\sum_h w_{ih}$.
{\sc Node2Vec} uses two hyperparameters $(p,q)$ to  control the "breadth" and "depth" of the walk, in particular, to what extent it remains in  the neighbourhood of the origin node.  The method initializes randomly the embedding vectors and updates them according to sentences.
Sentences for training are generated by selecting a start node uniformly. With both baseline methods, the distribution over sentences is {\em static}, that is remains the same in the course of training. To streamline the presentation, we will use the baselines as black boxes that take an input graph $G=(V,E,w)$ and for a node $i\in V$ and length $t>0$ provide samples from the baseline-specific distribution
$\Walks[i,t]$.

\subsection{Overview of SGNS} \label{SGNS:sec}
For completeness, we provide more details on SGNS \cite{Mikolov:NIPS13}.
SGNS trains two vectors for each entity $i$, a focus vector, $\boldsymbol{f}_i$, and a context vector $\boldsymbol{c}_i$.

SGNS takes as hyper parameters a "skip window" $\Delta$ and ratio $\lambda$ of positive to negative examples.  It works with input sentence $S:= (v_0,\ldots,v_t)$ as input.  A sentence $S$ is processed by generating a randomized set of pairs that are then used as positive training examples:
\begin{align*}
\text{(i)}&\, \text{Draw i.i.d.\ $\Delta_i \sim U[t]$ for $i\in [t]$}\\
\text{(ii)}&\, \Pairs(S) \gets \bigcup_{i=0}^t \left\{ (v_i,v_j) \mid |i-j|\leq \Delta_i\right\}\enspace .
\end{align*}
Skip lengths $\Delta_i$ for each $v_i$ are selected independently uniformly at random from $\{1,\ldots,t\}$. $\Pairs(S)$ then includes all ordered pairs where $v_j$ within that skip length from $v_i$.

For each positive example, $\lambda$ random negative examples are drawn with the same focus $i$ and a randomly selected context according to entity frequencies in positive examples to the power of $0.75$. 
Intuitively, negative examples \cite{HuKorenV:2008} provide an "anti-gravity" effect that prevents all embeddings from collapsing into the same vector.
We denote by $\kappa_{ij}$  the probability that positive example pair $(i,j)$ is generated and by 
 $n_{ij} \propto \|\kappa_{i\cdot}\|_1 \|\kappa_{\cdot j}\|_1^{0.75}$  the probability that a negative example pair $(i,j)$ is generated.  The hyper parameter $\lambda$ specifies a ratio of negative to positive examples.  The optimization 
objective when using this distribution over examples has the general form:
\begin{equation} \label{sgns_loss}
L := \sum_{i,j}\kappa_{ij}L_{+}(i, j) + \lambda \sum_{i,j}n_{ij}L_{-}(i, j)\enspace .
\end{equation}

The per-example loss functions are defined as
\begin{eqnarray}
L_{+}(i, j) & :=& \log(\sigma(\boldsymbol{f}_{i}\cdot \boldsymbol{c}_{j})) \\
L_{-}(i, j) & := & \log(\sigma(-\boldsymbol{f}_{i} \cdot \boldsymbol{c}_{j}))\enspace ,
\end{eqnarray}
where the sigmoid function is defined to be $\sigma(x) := \frac{1}{1+\exp(x)}$.

At a high level, the gradient updates on positive example $(i,j)$ increase the inner product $\boldsymbol{f}_i\cdot \boldsymbol{c}_j$ and an update on a negative example $(i,j)$ decreases that inner product.  The SGNS objective is designed to maximize the log likelihood over all examples.  This when
the probability of positive example $(i,j)$ is 
modeled by a sigmoid of the inner product
$\sigma(\boldsymbol{f}_{i}\cdot \boldsymbol{c}_{j})$ and that of a 
negative example by a sigmoid of the negated product
$\sigma(-\boldsymbol{f}_{i}\cdot \boldsymbol{c}_{j})$.
 The logarithm of the likelihood function has the form \eqref{sgns_loss}.

 To streamline the presentation, we will treat the SGNS as a closed module for computing embedding vectors $\{\boldsymbol{f}_i,\boldsymbol{c}_i\}$ for $i\in V$.
 The module inputs $V$, length parameter $t$, and window size $\Delta$.  It has a procedure to initializes the embedding vectors.  It then enters a training phase that takes as input {\em sentences} $S\in V^t$ and updates the embedding vectors.

\section{Loss-guided training methods} \label{lossguidedalgs:sec}
We first discuss the challenges and design goals for loss-guided training in the SGNS-based node embedding domain. 
Methods in prior work were designed for supervised learning, where examples are labeled.  In our setting, 
the SGNS loss (Equation~\ref{sgns_loss}) has both positive examples (that are generated from pairs co-occurring in random walks) and negative examples (that are selected randomly according to the distribution of positive examples). The negative examples distribution is therefore determined by the positive example distribution. Hence, in our setting the knob we modify would only be the distribution of positive examples.

Most methods in prior work compute (or track) approximate loss values for all examples.  In our setting, the set of potential positive examples is very large, can be quadratic in the representation of the input graph dataset, and these examples are generated rather than provided explicitly. Therefore, having to maintain even approximate loss values for all potential positive examples is not feasible and can severely impact efficiency.  We will instead aim to draw subsets of examples and select from these subsets according to current loss values.

Finally, the baseline methods we build on do not work with examples individually but instead generate random walks and multiple examples $\Pairs(S)$ from each walk $S$.
Using random walks rather than individual edges proved to be hugely beneficial and we do not want to lose that advantage in our loss-guided methods.  Therefore, our loss-guided selection methods stick to the paradigm of generating random walks $S$ and training with $\Pairs(S)$. 

\subsection{Loss-guided random walks}
Perhaps the most natural method to consider is to incorporate the loss values of edges in the random walks.
As in the baseline methods, the start node $v_0 \sim U[|V|]$ is selected uniformly at random.  A walk $(v_0,\ldots,v_t)$ of length $t$ is then computed so that

\begin{equation}
\Pr(v_{i}\mid v_{i-1}) = \frac{w_{v_{i-1},v_i} L_{+}(v_{i-1}, v_{i})^p}{ \sum_{ u \mid (v_{i-1},u)\in E \} } w_{v_{i-1},u}L_{+}(v_{i-1}, u)^p } \enspace ,
\end{equation}
where $p\geq 0$ is a hyper-parameter that tunes the dependence on loss.  A choice of $p=0$ provides the basic random walks used in ${\sc DeepWalk}$~\cite{deepwalk:KDD2014}. A large value of $p$ will result in always selecting the highest-loss outgoing edge.  A value of $p=1$ will select an edge proportionally to the product of its weight and loss value.
A drawback of this method is that it is less efficient computationally: When the random walk distribution is static we can preprocess the graph so that walk generation is very efficient.  Here we need to recompute loss values and edge probabilities of all outgoing edges while generating the walk. 
We observed empirically, however, that its performance in terms of training computation (per number of walks used for training) on almost all datasets is generally unstable and poor.  This prompted us to consider instead {\em loss-guided selection} of walks, where the candidate random walks for training are generated as in the baseline method but the selection of walks is made according to assigned {\em loss scores}. 

\subsection{Loss-score based selection of walks}

 \SetKwFunction{Traininit}{Train.Initialize}
 \SetKwFunction{Trainupdate}{Train.Update}
 \SetKwFunction{Train}{Train}
 We propose a design that addresses the general issues and those specific to our settings.  At a high level, we use the same random walk distribution and update and training procedures as the baseline methods (see Algorithm~\ref{alg:baseline}) but we modify the selection of walks for training.
 Algorithm~\ref{alg:walkscore} is a meta-algorithm for our loss-guided walk selection methods.  The pseudo-code treats components as "black-boxes:" (i)~The random walk distribution $\Walks[i,t]$  generated from a graph $G=(V,E,w)$ according to a random process, specified start node $i\in V$ and specified length $t$.  
 (ii)~A training algorithm $\Train$ (such as a variant of SGNS) that includes an initialization method $\Train.{\sc Initialize}$ of the embedding vectors $\{\boldsymbol{f}_i,\boldsymbol{c}_i \}$ and an update method $\Train.{\sc Update}$  that inputs sentences (walks) $S$, generates from them positive and negative examples (according to parameters on example generation $\Delta$ and $\lambda$), and performs the respective parameter updates.  
  A component that is used only with the loss-guided methods is a
  {\em loss scoring} function $\Lscore(S)$ of walks. Our choice of functions will be detailed later but they depend on specified power $p>0$ and may also depend on the specifics of example generation from walks (see $\Pairs(S)$ in Section~\ref{SGNS:sec}).
    
 For training, we initialize the embedding vectors and then repeat the following {\em rounds}: We draw random walks $S_i\sim \Walks[i,t]$, one generated for each node $i\in V$, we compute loss scores $\Lscore(S_i)$ for each of those $|V|$ walks.  We then select a subset of these walks for training in a way that is biased towards the walks with the higher loss score.   Specifically we will use an integer parameter $F\geq 1$ and select for training $|V|/F$ of the scored walks.  The selection within each round is done using a weighted sampling without replacement method according to the loss scores $\Lscore(S_i)$ of the walks $i\in V$.  The weighted sampling can be implemented very efficiently in a single distributed pass over walks using each one of a variety of known order/bottom-$k$/varopt sampling methods (e.g.~example~\cite{Rosen1997a,Ohlsson_SPS:1998,Cha82,bottomk07:ds,bottomk07:est,DLT:jacm07,varopt_full:CDKLT10}).
 Finally, the selected walks from the round are handed to the training algorithm. 
 
 The meta procedure selects in each of \#epochs$*F$ {\em rounds} a set of $|V|/F$ walks.  Therefore, selecting a total of \#epochs$*|V|$ walks in total for training.
 In order to compare with the baseline methods that select $|V|$ walks per epoch (one from each node), we use the term epoch to refer to providing $|V|$ walks for training. The pseudocode lists parameters that are used in various
  "black box" components: The length $t$ of the generated walks, window size $\Delta$ used to generate positive examples $\Pairs(S)$ from a walk $S$, and a power $p>0$ which we will use later as a parameter in the scoring of walks are passed to the respective components.

 \begin{algorithm2e}[htbp]\caption{Baseline method
   \label{alg:baseline}}
\DontPrintSemicolon
 \KwIn{Graph $G=(V,E,w)$; Random Walk dist  $\Walks[i,t]$ over $V^t$;
 walk length $t$;  \#epochs; Training method $\Train$ (that uses window $\Delta$ and negatives ratio $\lambda$)}
 $\Traininit(V)$ \tcp*{Initialize embedding parameters $\{\boldsymbol{f}_i,\boldsymbol{c}_i\}$ for ${i\in V}$}
 
   \ForEach(\tcp*[h]{Select walks and update}){$r\in [\text{\#epochs}]$} 
   {
   \ForEach{$v\in V$ (shuffled)}{
   Draw $S_v \sim \Walks[v,t]$\;
   $\Trainupdate(S_v)$ \tcp*{Train on Walk $S_v$ with window size $\Delta$}
   }
   }
   \Return{Embedding vectors $\{\boldsymbol{f}_i,\boldsymbol{c}_i\}$ for ${i\in V}$}
\end{algorithm2e}

\begin{algorithm2e}[htbp]\caption{Walk selection by loss score
   \label{alg:walkscore}}
\DontPrintSemicolon
 \KwIn{Graph $G=(V,E,w)$; Random Walk dist  $\Walks[i,t]$ over $V^t$;
 walk length $t$; \#epochs; $F$ (fraction of walks selected per round); Walk scoring function $\Lscore$ (that uses power $p\geq 0$); Training method $\Train$ (that uses window $\Delta$ and negatives ratio $\lambda$)}
 $\Traininit(V)$ \tcp*{Initialize embedding parameters $\{\boldsymbol{f}_i,\boldsymbol{c}_i\}$ for ${i\in V}$} 
   \ForEach(\tcp*[h]{training rounds}){$\textit{round} \in [\text{\#epochs} * F]$}
   {
   \ForEach(\tcp*[h]{Draw and score $|V|$ walks}){$v\in V$}{
   Draw $S_v \sim \Walks[v,t]$\;
   $L_v \gets \Lscore(S_v)$
   }
   $D\gets $ A weighted sample without replacement of size $|V|/F$ from $V$ according to weights $L_v$\;
      \ForEach(\tcp*[h]{Train on walks $D$}){$v\in D$ (shuffled)}
   {
   $\Trainupdate(S_v)$ 
           }
   }
   \Return{Embedding vectors $\{\boldsymbol{f}_i,\boldsymbol{c}_i\}$ for ${i\in V}$}
\end{algorithm2e}

 \subsection{Loss scoring of walks} \label{lossscoring:sec}
 We consider several ways to assign loss scores to a walk $S := (v_1,\ldots, v_t)$ and respective $\Pairs(S)$.  All methods use a power parameter $p$.
 Our first scoring function uses the average loss of all positive examples generated from walk $S$:
\begin{equation}\label{Lscore1}
 \textstyle{\Lscore}_{\text{all}}(S) := \sum_{(i,j)\in \Pairs(S)} L_+(i,j)^p\enspace .
\end{equation}
  The second function heuristically scores a walk by its first $t'\in [t]$ edges
 \begin{equation}\label{Lscore3}
 \textstyle{\Lscore_{t'}}(S) := \sum_{i=1}^{t'-1} L_+(v_i,v_{i+1})^p\enspace .
  \end{equation}
With $t'=1$, the walk is scored by its first edge 
 $\textstyle{\Lscore_1}(S) := L_+(v_1,v_2)^p$.

 The advantage of the loss score $\Lscore_{t'}$ over $\Lscore_{\text{all}}$ is that we can compute the loss scores for a candidate walk $S_i$ from a node $i$ without explicitly computing the walk: It suffices to draw only the first $t'$ edges $S_{i1},\ldots,S_{it'}$ of a walk. If the node $i$ is selected to the sample $D$, only then we can sample the remaining edges of the walk $S_i$, conditioned on its prefix being $S_{i1},\ldots,S_{it'}$.   Since we compute loss scores to $F$ times many walks than we actually train with, this is considerable saving in our per-round preprocessing cost.
 The disadvantage of the loss score~\eqref{Lscore3} is that we are only using $t'$ examples from the set $\Pairs(S)$ to determine the loss score, so we can expect a reduction in effectiveness.
 
 The power $p$ in the computation of loss scores has the role of a hyperparameter:  High values of $p$ focus the selection more on walks with high loss examples whereas lower values allow for a broader representation of walks in the training.
 Interestingly,  Shalev-Shwartz and Wexler~\cite{shalev:ICML2016} considered the more extreme objective of minimizing the maximum per-example loss.  This objective is very sensitive to outliers (persistent high loss examples) and in some cases can divert all the training effort to be futilely spent on erroneous examples.   Note that in our setting, we are not as exposed because the walks pool we select from in each round is randomized and we use without replacement sample to select a $1/F$ fraction of that pool for training.
 
 \subsection{Complexity analysis} \label{complexity:sec}
 
 The per-epoch training cost with both the baseline and loss-guided selection methods amounts to computing the gradients
 of the loss functions $L_+(i,j)$ or $L_{-}(i,j)$ (for positive and negative examples) and applying gradient updates.
The training cost is proportional to the total number of examples generated from 
 $|V|$ walks.  The expected number of positive examples, $\E[|\Pairs(S)|]$, depends on the walk length $t$ and window $\Delta$.  The total number also depends on the  negatives to positives ratio $\lambda$ (see Section~\ref{SGNS:sec}). Therefore, each walk $S$ generates in expectation $\E[|\Pairs(S)|] (\lambda+1)$ training examples.  We train on $|V|$ walks in each epoch and thus the per-epoch training cost is:
 \begin{equation}
     C_{\text{train}} := \E[|\Pairs(S)|] (\lambda+1) |V| \ .
 \end{equation}
 
 We next consider the per-epoch total computation cost, that includes preprocessing cost.  
  For the baseline methods, the
 preprocessing cost 
  corresponds to generating $|V|$ random walks ($t$ edge traversals each\notinproc{\footnote{Node2Vec requires keeping large state for efficient walk generation, but this will not affect much our comparative analysis of baseline versus loss-guided methods.}}).  The total cost is dominated by the gradient computations of the training cost and is:
 \begin{equation}
      C_{\text{baseline}} = |V|  (\lambda+1) \E[|\Pairs(S)|] \enspace .
\end{equation}      
  
  For the loss-guided methods, the preprocessing cost involves evaluations of the loss $L_+(i,j)$ on positive examples. 
  With loss score $\Lscore_{t'}$ in each round we generate the first $t'$ steps of a random walk from each node $i\in V$. We then evaluate the loss score for each of the walks, which amounts to evaluating $L_+(i,j)$ on $t'$ pairs (only $|V|/F$ of the walks are selected for training). The total number of loss evaluation per epoch ($F$ rounds) is $C_{\text{prep,$t'$}} = F |V| t'$.
   With the loss score $\Lscore_{\text{all}}$ we generate in each round a complete walk from each node and evaluate the $L_+(i,j) $ for each pair in $\Pairs(S)$.  The total number of loss evaluations is 
 $C_{\text{prep,all}} := \E[|\Pairs(S)|] F |V|$. 
The total computation cost combines the training and preprocessing cost and is measured by the number of loss or gradient evaluations. Note that loss or gradient evaluations have similar complexity and amount to computing $e^{<x,y>}$ for loss and $x e^{<x,y>}$ for the gradient. 
    Summarizing, we have
    {\small
  \begin{align}
     C_{\text{$t'$}} &= |V|F t' + (\lambda+1) \E[|\Pairs(S)|] =  |V|F t' + C_{\text{baseline}}\\
     C_{\text{all}} &= |V|(F +\lambda+1) ( \E[|\Pairs(S)|] = |V| F\E[|\Pairs(S)|]+ C_{\text{baseline}}.
 \end{align}
    }

\section{Empirical Evaluation Setup} \label{sec:empirical_evaluation}
As baseline and for walk generation with our methods we used
{\sc DeepWalk} \cite{deepwalk:KDD2014} and {\sc Node2Vec} \cite{node2vec:kdd2016}.  These methods define the random walk distributions $\Walks[i,t]$. 
When evaluating our methods we fit hyperparameters to the respective baseline method.  With
{\sc Node2Vec} we searched over values $p,q \in \{0.25, 0.5, 1, 2\}$.

We trained models using the Gensim package \cite{rehurek_lrec} that builds on a classic implementation of SGNS \cite{Mikolov:github2013}. We used the default parameters that generally perform well:  $t=10$ for the length of the walk (sentence), $\Delta=10$ for window size, and $\lambda=5$ for the number of negative examples generated for each positive example. With these values, we have in expectation $\E[|\Pairs(S)|]=63$ positive examples generated from each walk and $\E[|\Pairs(S)|](\lambda+1)=380$ examples in total generated for each walk when training.

 In our implementation we applied the baseline method (Algorithm~\ref{alg:baseline}) for the first epoch (training on $|V|$ walks $S_i\sim \Walks[i,t]$) and applied loss-guided methods (Algorithm~\ref{alg:walkscore}) starting from the second epoch.  This is because we expect scoring by loss to not be helpful initially, with random initialization.
 We used $F \in \{ 2, 5, 10, 20 \}$ rounds per epoch and power of the loss value $p \in \{1, 4, 32\}$. 
Each experiment is repeated $r$ times and we report the average quality and standard error. 
We fit parameters on one dataset from each collection using one set of repetitions and use the same parameters with all datasets and a fresh set of repetitions.

 As mentioned, SGNS determines the distribution of negative examples according to the frequencies of words in the provided positive examples. 
With the baseline methods, the distribution of random walks and hence the frequencies of words in positive examples remain fixed throughout training and are approximated by maintaining historic counts from the beginning of training.
 With our loss-guided selection, the distribution of positive examples changes over time.  We experimented with different variations that use a recent positive distribution (per-round or for few recent epochs) to guide the negative selection.  We did not observe significant effect on performance and report results with respect to frequencies collected since the start of training.

\subsection{Tasks and metrics} \label{sec:metrics}
We evaluated the quality of the embeddings on the following tasks, using corresponding quality  metrics:
\begin{trivlist}
  \item[$\bullet$]
{\em Clustering:}  The goal is to partition the nodes into $k$ clusters. The embedding vectors are used to compute a $k$-means clustering of nodes. We used \texttt{sklearn.cluster.KMeans} from \texttt{scikit-learn} package~\cite{scikit-learn} with default parameters. Our quality measure is the
 {\em modularity} score  \cite{Modularity:PhysRev2004} of the clustering.
\item[$\bullet$]
{\em Multi-class (or multi-label) classification:} Nodes have associated classes (or labels) from a set $L$.  The class (or the set of labels) are provided for some nodes  and the goal is to learn the class/labels of remaining nodes. 
An embedding is computed for all nodes (in an unsupervised fashion).  Following that, a supervised learning algorithm is trained on embedding and class/label pairs.
We used One-vs-Rest logistic regression from the \texttt{scikit-learn} package\notinproc{\footnote{\texttt{sklearn.multiclass.OneVsRestClassifier}}} with default parameters \cite{scikit-learn}.  For multi-labels we used the \texttt{multinomial} option.  
In a multi-class setting, we obtain a class prediction from the embedding vector for each of the remaining nodes and report the fraction of correct predictions.
In a multi-label setting, we provide the number of labels and the embedding vector and obtain a set of predicted labels for each node.  
We report the micro-averaged F1 score. 
\end{trivlist}

\subsection{Measuring gain} \label{sec:gain}
 Across our datasets, peak accuracy with loss-guided selection was equal or higher than baseline.  We thus consider efficiency, which we measure using $\nepochs(\text{method})$, the average number of training epochs 
 over repetitions
needed for the method to reach $0.95$ of peak accuracy.
We can now express the training and computation cost and respective gains.
With the parameter values we use, 
the per-epoch training cost is
$C_{\text{train}}= 380 |V|$ and the per-epoch computation cost is
$C_{t'} = |V| (Ft' + 380)$  and $C_{\text{all}} = |V|(F+6)63$.
Accordingly, we express the gain of a loss-guided method
with scoring function  $\Lscore_{\text{method}}$
with respect to the baseline:
\begin{trivlist}
\item[$\bullet$]
{\em Training gain:} is the relative decrease in number of training epochs (recall that training cost per epoch is similar for all methods).
\begin{equation} \label{traingain:eq}
    1-\frac{\nepochs(\text{method})}{ \nepochs(\text{baseline})}\enspace .
\end{equation}
When reporting the training gain, we report the error over repetitions: We compute the (sample) standard deviation of the number of epochs used by the method to reach peak (over repetitions) and normalize it by dividing by $\nepochs(\text{baseline})$.
\item[$\bullet$]
{\em Computation gain:} The relative decrease in computation cost
\begin{equation} \label{compgain:eq}
1-\frac{C_{\text{method}}\cdot \nepochs(\text{method})}{C_{\text{train}}\cdot\nepochs(\text{baseline}) }\enspace .
\end{equation}
With $\Lscore_{t'}$ the computation gain is $1- \frac{(380+Ft')\nepochs(t')}{380\nepochs(\text{baseline})}$ and with $\Lscore_{\text{all}}$ it is $1- \frac{(380+63F)\nepochs(\text{all})}{380\nepochs(\text{baseline})}$. 
\end{trivlist}

\section{Datasets and tasks} \label{datasets:sec}
We evaluate our methods on three collections of real-world datasets, summarized in Table~\ref{tab:datasets}. The datasets have different learning tasks (see Section~\ref{sec:metrics}):
\begin{trivlist}
    \item[$\bullet$] {\em Facebook page networks (clustering)}: The collection represent mutual "like" networks among verified Facebook pages. There are six datasets for different communities (TV shows, athletes, and more)~\cite{DBLP:conf/asunam/RozemberczkiDSS19}. 
    The task (following~\cite{DBLP:conf/asunam/RozemberczkiDSS19}) is to compute embeddings with $d=16$ and cluster the data to $k=20$ clusters.  .
    
    \item[$\bullet$] {\em Citation networks (multi-class)}: The collection has three networks (Citeseer, Cora and Pubmed) \cite{DBLP:journals/aim/SenNBGGE08}. Networks are formed by having a node for each document and an (undirected, unweighted) edge for each citation link. Each document has a class label. Following~\cite{node2vec:kdd2016,Yang:ICML2016}, we train a $d=128$ dimensional embedding and use a random selection of 20 nodes per class as labeled training examples. 

    \item[$\bullet$] {\em Protein-Protein Interactions (PPI) (multi-label)}: The dataset is a graph of human Protein-Protein interactions \cite{node2vec:kdd2016}.  Each protein (node) has multiple labels and the goal is to predict this set of labels.
    Following \cite{node2vec:kdd2016},  we use $d=128$ and use 50\% of nodes (selected uniformly at randomly) for training. 
\end{trivlist}

\begin{table}[]
    \centering
    \footnotesize
    \begin{tabular}{c|c|c|c}
        \hline
  dataset  & $|V|$ & $|E|$ & \\
  \hline\hline
  \multicolumn{3}{c}{\cite{DBLP:conf/asunam/RozemberczkiDSS19}  Facebook pages} & clustering $k$\\
  \hline
        Athletes & $13,866$ & $86,859$ & 20 \\
        Company & $14,113$ & $52,311$ & 20 \\
        Government & $7,057$ & $89,456$ & 20 \\
        New Sites & $27,917$ & $206,259$ & 20 \\
        Politicians & $5,908$ & $41,730$ & 20 \\
        Public Figures & $11,565$ & $67,115$ & 20 \\
        TV Shows & $3,892$ & $17,263$ & 20 \\
\hline
\multicolumn{3}{c}{\cite{DBLP:journals/aim/SenNBGGE08} Citation networks} & Multi-class $|L|$ \\
\hline
        Pubmed & $19,717$ & $44,338$ & $3$ \\
        Cora & $2,708$ & $5,429$& $7$ \\
        Citeseer & $3,327$ & $4,732$ & $6$\\
\hline
\multicolumn{3}{c}{\cite{node2vec:kdd2016}  Protein Interactions} & Multi-label $|L|$ \\
\hline
       PPI & $3890$ & $38739$ & $50$ \\
    \end{tabular}
    \caption{Summary of dataset properties and tasks.}
    \label{tab:datasets}
\end{table}

\section{Empirical Results} \label{results:sec}

We evaluate our methods using three key metrics: Quality, training gain, and computation gain.  We use figures to show quality in the course of training: We plot  average performance over repetitions and provide error bars that correspond to one SD. We use tables to report training and computation gains for different methods and hyper-parameter settings.  
In Appendix~\ref{param_sensitivity:sec} we provide parameter sweeps on the number of rounds per epoch $F$, the loss power $p$, and $t'$ in the loss score $\Lscore_{t'}$.
In this section we report results for $F=10$ rounds per epoch, which seems to be a sweet spot for the training cost (see Appendix~\ref{param_sensitivity:sec} for other values of $F$). We use both {\sc DeepWalk} and {\sc Node2Vec} baselines\notinproc{ (Additional results reported in Appendix~\ref{moren2v:sec})}. For each loss scoring function we used the best performing overall power $p$: $\Lscore_1$ performed well with $p=32$ (selecting the $|V|/10$ highest loss walks in each round).  $\Lscore_{10}$ performed well with $p=4$ (weighted sampling that is biased towards higher loss). 
Interestingly, $\Lscore_{all}$ did not perform better than $\Lscore_{10}$ even in terms of training cost (since it is computation heavy, there is also no improvement in computation cost).  We show performance with $\Lscore_{all}$ in plots but do not report it in tables.\notinproc{ Appendix~\ref{loss:sec} provides additional exploration on the loss patterns of loss-guided versus baseline training.}

\subsection{Facebook networks (clustering task)}

\begin{figure*}[ht]
\vskip 0.2in
\begin{center}
\includegraphics[width=0.65\columnwidth]{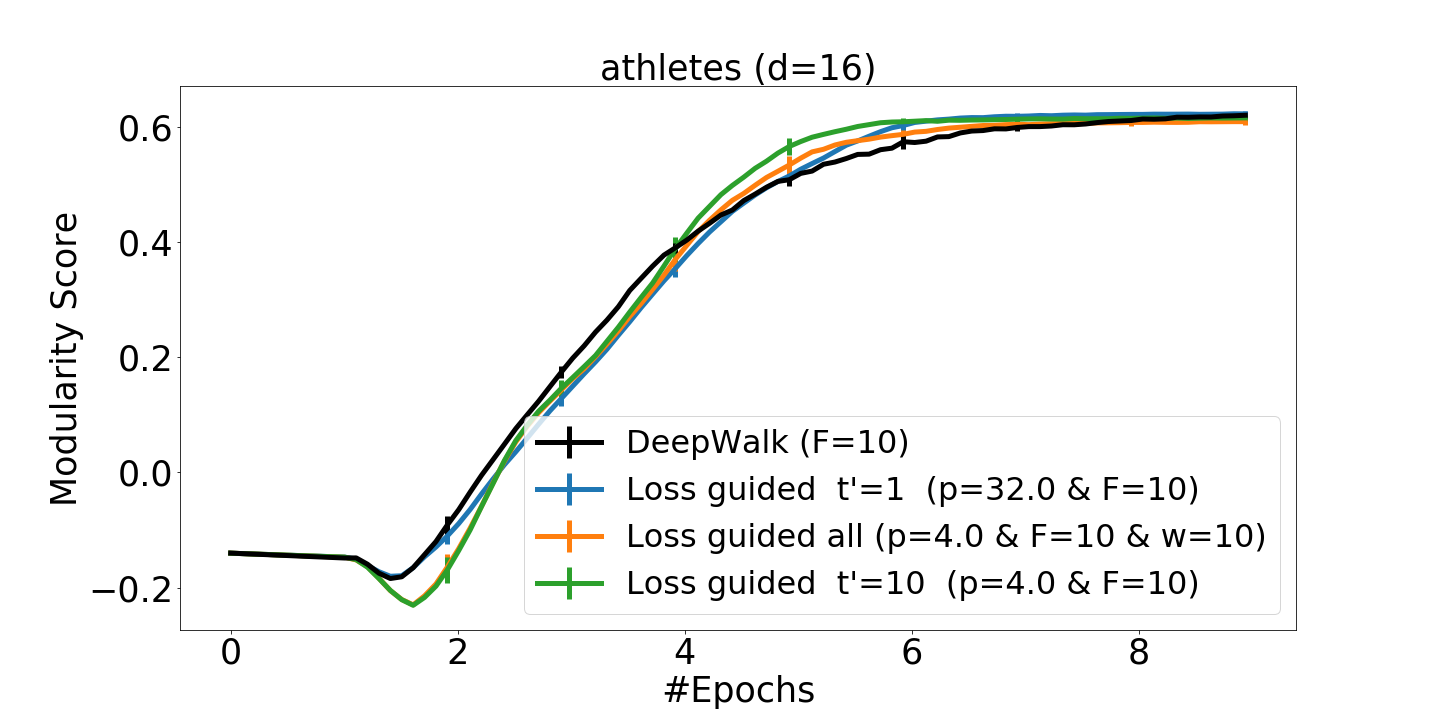}
 \includegraphics[width=0.65\columnwidth]{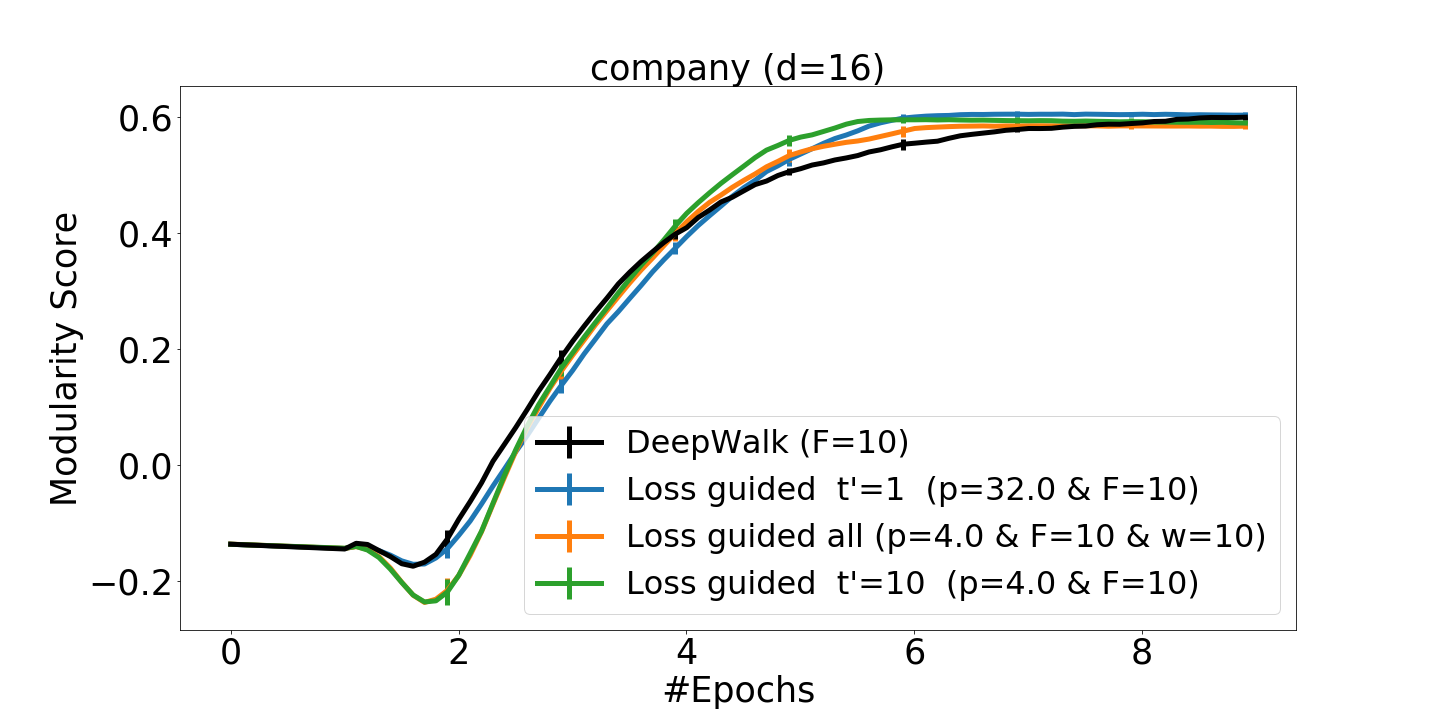}
\includegraphics[width=0.65\columnwidth]{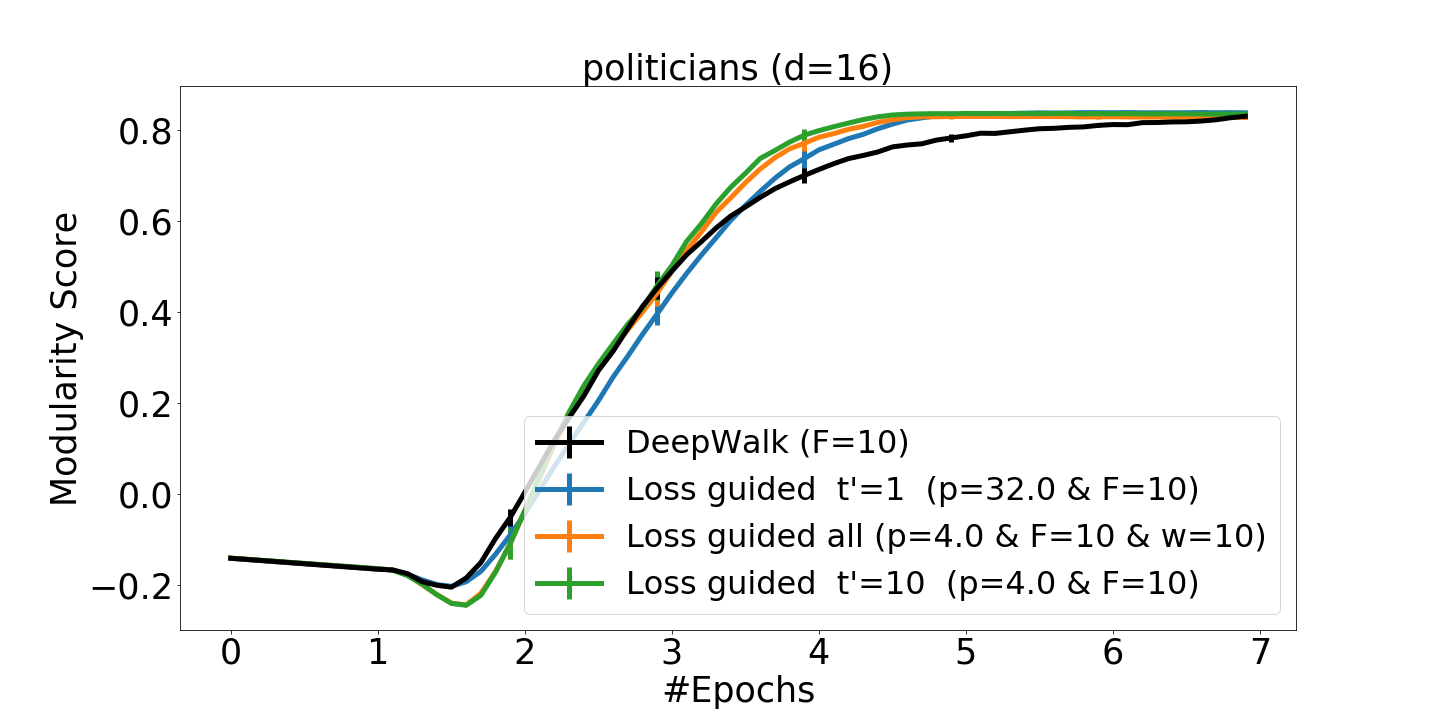}
\caption{Modularity score in the course of training on the Athletes, Company, and Politicians networks from the Facebook collection. We show {\sc DeepWalk} and loss-guided with different walk scoring functions.}
\label{plot:dynamic_gemsec}
\end{center}
\vskip -0.2in
\end{figure*}

Representative results based on $r=200$ repetitions are reported in 
Table~\ref{tab:facebook_network_optimal_params_performance} for both baselines.
Figure~\ref{plot:dynamic_gemsec} shows the 
modularity score in the course of training for representative datasets and methods.
We fitted the Node2vec parameters to $p=2$ and $q=1$ on the Athletes datasest and applied with all datasets and methods.
We see that loss-based selection obtained 13\%-25\% reduction in training and 6\%-20\% reduction in computation for both baselines.
We can see that on almost all datasets in this collection $\Lscore_{10}$ outperformed $\Lscore_1$ in terms of training cost but in most cases had a lower overall gain in computation cost.

\begin{table}[]
    \centering
     \footnotesize
    \begin{tabular}{c|c|c c|c||c c | c}
       &   & \multicolumn{2}{c}{Training } & Comp  & \multicolumn{2}{c}{Training } & Comp\\
        dataset&$t'$, $p$ &  \%gain & \%SD & \%gain & \%gain & \%SD & \%gain \\
        \hline
        \multicolumn{5}{c}{{\sc DeepWalk} baseline} & \multicolumn{3}{c}{{\sc Node2Vec} baseline}\\
        \hline
        Athletes & $1$, $32$ & 12 & 1.8 & {\bf 9.8} & 12.91 & 2.40 & {\bf 10.7} \\
          & $10$, $4$ & {\bf 18.2} & 3.10 &0.70 & {\bf 18.2} & 2.33 &0.14 \\
        \hline
        Company & $1$, $32$ & 18.2 & 1.86 & {\bf 16.0} & 20.0 & 2.21 & {\bf 17.7}\\
         & $10$, $4$ & {\bf 22.3} &1.71 & 5.5 & {\bf 22.6} &1.50 & 5.38\\
        \hline
        Government & $1$, $32$ & 10.7 & 2.67 & {\bf 8.47}& 10.4 & 2.10 & {\bf 8.13}\\
         & $10$, $4$ &{\bf 21.9} & 1.90 & 5.60 &{\bf 20.3} & 2.10 & 2.61\\
        \hline
          New Sites& $1$, $32$ & {\bf 15.2} & 3.5 & {\bf 10.1} & {\bf 12.5} & 3.63 & {\bf 10.5} \\
          & $10$, $4$ &4.21& 9.58 & -17.7 &7.08& 8.70 & -14.5\\
    
        \hline
        Politicians &  $1$, $32$  & 17.9 & 2.19 & {\bf 15.8} & 18.2 & 2.51 & {\bf 16.0}\\
          & $10$, $4$ & {\bf 24.6} & 1.60 & 9.44  & {\bf 24.1} & 1.73 & 8.16\\
        \hline
        Public & $1$, $32$ & 10.2 & 5.08 & 7.93 & 7.10 & 3.81 & 4.9\\
        Figures & $10$, $4$ &{\bf 24.3} & 2.97 & {\bf 7.69 } &{\bf 23.3} & 1.84 & {\bf 6.05 }\\
        \hline
        TV Shows & $1$, $32$ &21.76 & 1.14 & {\bf 19.65} &21.6 & 1.63 & {\bf 20.0}\\
           & $10$, $4$ & {\bf 25.2} & 1.51 & 9.88& {\bf 24.4} & 1.3 & 8.88\\
\hline
    \end{tabular}
    \caption{Facebook Pages: Training and computation gain of loss-guided with
    {\sc DeepWalk} and {\sc Node2Vec} ($(p,q)=(2,1)$) baselines ($r=200$ repetitions, $F=10$ rounds per epoch, $(\Lscore_1, p=32)$ and $(\Lscore_{10},p=4)$)}
    \label{tab:facebook_network_optimal_params_performance}
\end{table}

\begin{figure}[ht]
\vskip -0.8in
\begin{center}
\centerline{
\includegraphics[width=\columnwidth]{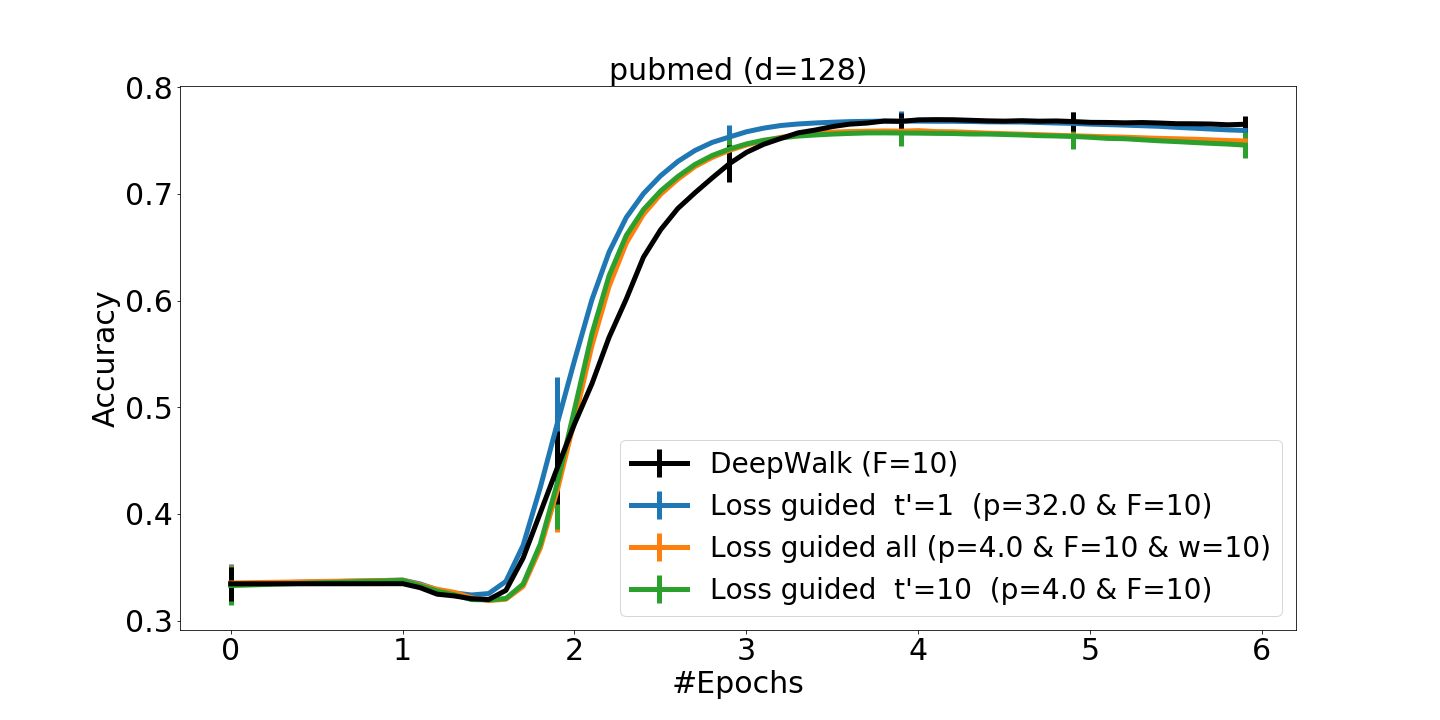}
}
\caption{Pubmed dataset: Accuracy in the course of training for {\sc DeepWalk} and loss-guided methods with different scoring functions. 
}
\label{plot:citations}
\end{center}
\vskip -0.4in
\end{figure}

\begin{figure}[ht]
\vskip 0.2in
\begin{center}
\centerline{
\includegraphics[width=\columnwidth]{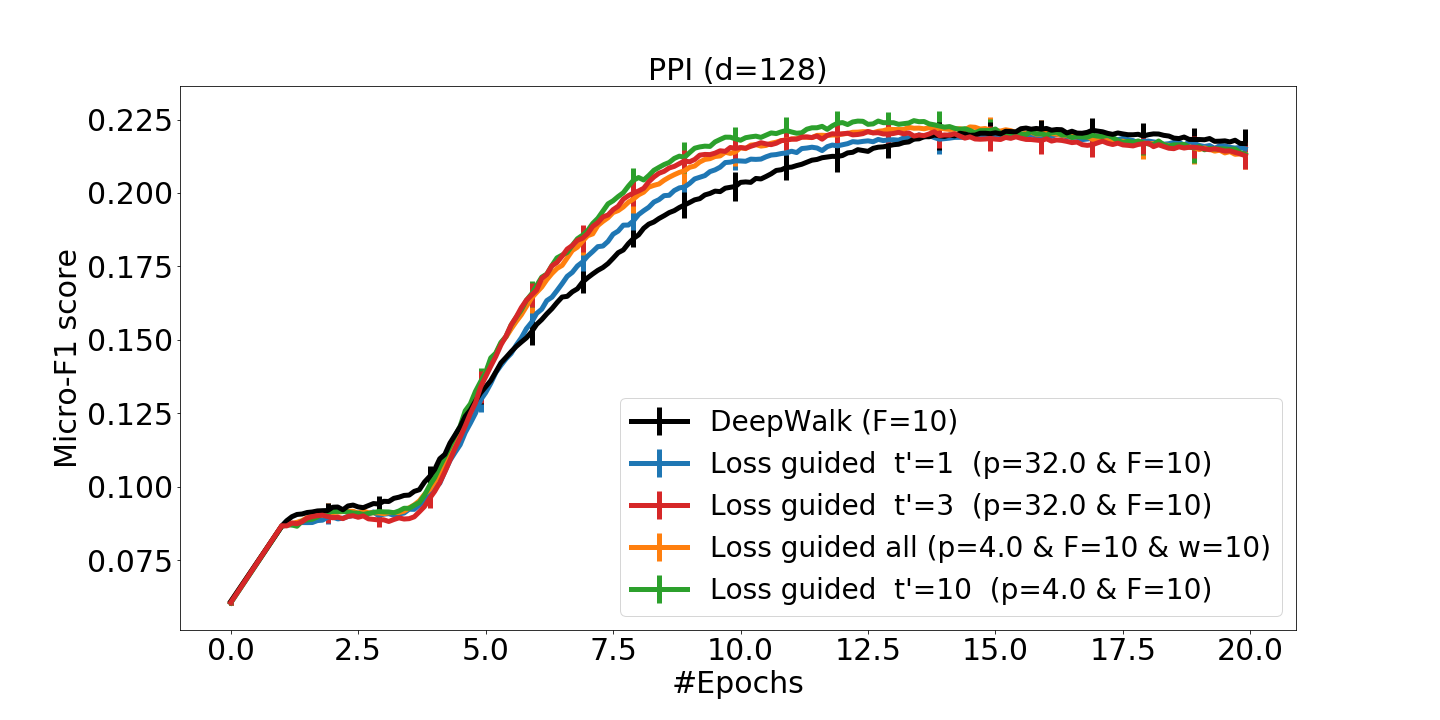}
}
\caption{PPI dataset: Averaged micro-F1 score in the course of training for {\sc DeepWalk} and loss-guided methods with different scoring functions.
}
\label{plot:ppi}
\end{center}
\vskip -0.2in
\end{figure}

\subsection{Citation networks (Multi-class)}
Representative results with $r=400$ repetitions are reported in 
Table~\ref{tab:citation_network_optimal_params}. Figure~\ref{plot:citations} shows performance in the course of training for the Pubmed dataset. 
The {\sc node2vec} parameters were fitted on the Cora dataset to $(p,q)=(2,0.25)$.
We can see that the loss-guided methods had training gains of 8\%-12\% on the Pubmed and Citeseer datasets, but due to large error bars there is no significance for the improvements on Cora.  
The loss score $\Lscore_1$ outperformed others also in terms of training cost.

\begin{table}[]
    \centering
     \footnotesize
    \begin{tabular}{c|c|c c|c||c c | c}
       &   & \multicolumn{2}{c}{Training } & Comp  & \multicolumn{2}{c}{Training } & Comp\\
        dataset&$t'$, $p$ &  \%gain & \%SD & \%gain & \%gain & \%SD & \%gain \\
        \hline
        \multicolumn{5}{c|}{{\sc DeepWalk} baseline} & \multicolumn{3}{c}{{\sc Node2Vec} baseline}\\
         \hline
         \multicolumn{8}{c}{Citation Networks, $r=400$}\\
        \hline
        Pubmed & $1$, $32$ & {\bf 9.07} & 3.91 & {\bf 7.38} & {\bf 9.06} & 3.55 & {\bf 8.28}  \\
         & $10$, $4$ &  5.21 & 7.02 &-10.5 &  6.14 & 5.96 &-10.7\\
        \hline
        Cora & $1$, $32$ & 1.80 & 8.60 & 0.00 & {\bf 4.08} & 7.45 & {\bf 4.23}\\
         & $10$, $4$ & {\bf 5.20} &8.12 & -12.4 & 8.27 & 6.24 & -9.84  \\
        \hline
        Citeseer & $1$, $32$ & {\bf 7.64} & 6.60 & {\bf 5.81} & {\bf 11.57} & 5.43 & {\bf 9.6}\\
         & $10$, $4$ & 5.73 & 6.20 & -11.2 & 7.90 & 8.37 & -9.90 \\
\hline\hline
\multicolumn{8}{c}{Protein Interaction Network, $r=200$}\\
\hline
PPI & $1$, $32$ & 12.7 & 3.91 &  3.90  & 10.4 & 7.82 &  10.7  \\
         & $3$, $32$ &  20.7 & 3.77 & {\bf 14.75} & 21.4 & 3.73 & {\bf 11.8} \\
         & $10$, $4$ & {\bf 22.2} & 3.38 & 4.06 & {\bf 22.2} & 3.90 & 4.50\\
        \hline
\end{tabular}  
    \caption{Citation and PPI Networks: Training and computation gain of loss-guided selection with
    {\sc DeepWalk} and {\sc Node2Vec} baselines. $F=10$, $(\Lscore_1, p=32)$, $(\Lscore_{10},p=4)$ and for the PPI network also $(\Lscore_3, p=32)$. }
    \label{tab:citation_network_optimal_params}
    \label{tab:ppi_network_optimal_params}
\vspace{-0.3in}    
\end{table}

\subsection{PPI network (multi-label)}
Representative results with $r=200$ repetitions with loss scores 
$\Lscore_{t'}$ for $t'\in\{1,3,10\}$
are reported in 
Table \ref{tab:ppi_network_optimal_params} and Figure~\ref{plot:ppi}.  
{\sc Node2vec} parameters were fitted to $(p,q)=(1,2)$. 
We observe that training costs improves with $t'$, and in particular the training gain with $\Lscore_{10}$ is  significantly higher than with $\Lscore_1$, but most of 
that gain is already attained by $\Lscore_3$. The computation gain is largest with $\Lscore_3$, which attains nearly the same training gain as $\Lscore_{10}$ but at lower per-epoch computation.  Overall, we see training gains of 22\% and computation gains of 12\%-15\%.

\section{CONCLUSION} \label{conclusion:sec}
We study loss-guided example selection, known to accelerate training in some domains, for
methods such as {\sc DeepWalk} and {\sc Node2Vec} that learn node embeddings using random walks. The random walk base methods use a static distribution over an implicitly-represented extended set of training examples and seems less amenable for dynamic loss-guided example selection.  We propose efficient methods that facilitate loss-based dynamic example selection while retaining the highly effective structure of random walks and scalability.  We demonstrate empirically the effectiveness of the proposed methods.  An interesting open question is to explore such benefits with other frameworks that generate training examples on-the-fly from an implicit representation such as example augmentation or together with methods that work with feature representation of nodes such as Pinsage~\cite{PinSage:KDD2018}.

\small{

\smallskip
\noindent
{\bf Acknowledgements}
This research is partially supported by the Israel Science Foundation (Grant No. 1595/19).  We thank the anonymous GRADES-NDA '20 reviewers for many helpful comments.}
\onlyinproc{\newpage}

\bibliography{main.bib}
\bibliographystyle{plain}

\appendix

\section{Synthetic communities graph} \label{case_study:sec}

\begin{figure}
\vspace{-0.2in}
    \includegraphics[width=0.4\columnwidth]{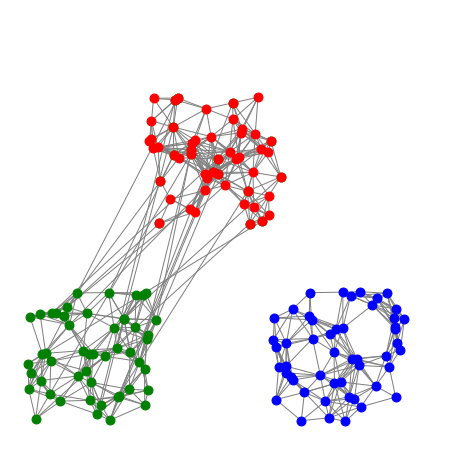}
    \caption{A community structure with three communities in green, red, and blue. The red and green communities are inter-connected and the blue community is isolated.}
    \label{plot:communities_illustration}
    \vspace{-0.1in}
\end{figure}

We start with a simple synthetic network that demonstrates the benefits of loss-guided selection. The example network structure is illustrated in Figure~\ref{plot:communities_illustration}.  We have three communities (red, green, and blue) of the same size. The red and green communities are inter-connected and the blue community is isolated. The goal is to reconstruct the ground truth community affiliations from the learned embedding. 
Our construction is inspired by random GnP graphs (each community is a GnP graph) and the planted partition model \cite{DBLP:journals/rsa/CondonK01}. Each of the communities has $10^4$ nodes.   We generated intra-community edges so that each pair $(i,j)$ of same-community nodes has a connecting edge with probability $p=0.001$.  Each inter-community pair $(i,j)$ from the red and green communities has a connecting edge with probability $q=0.0003$.  

\begin{figure}
    \includegraphics[width=0.45\columnwidth]{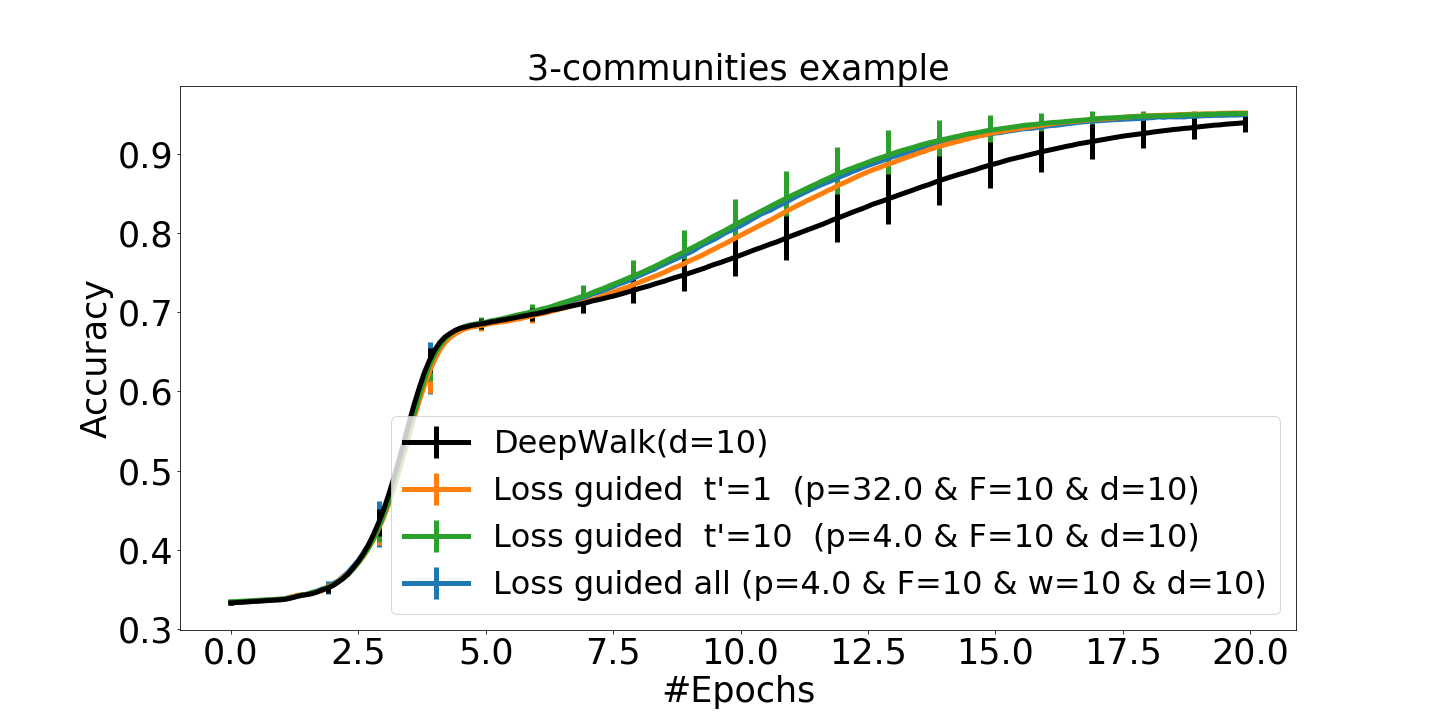}
    \includegraphics[width=0.45\columnwidth]{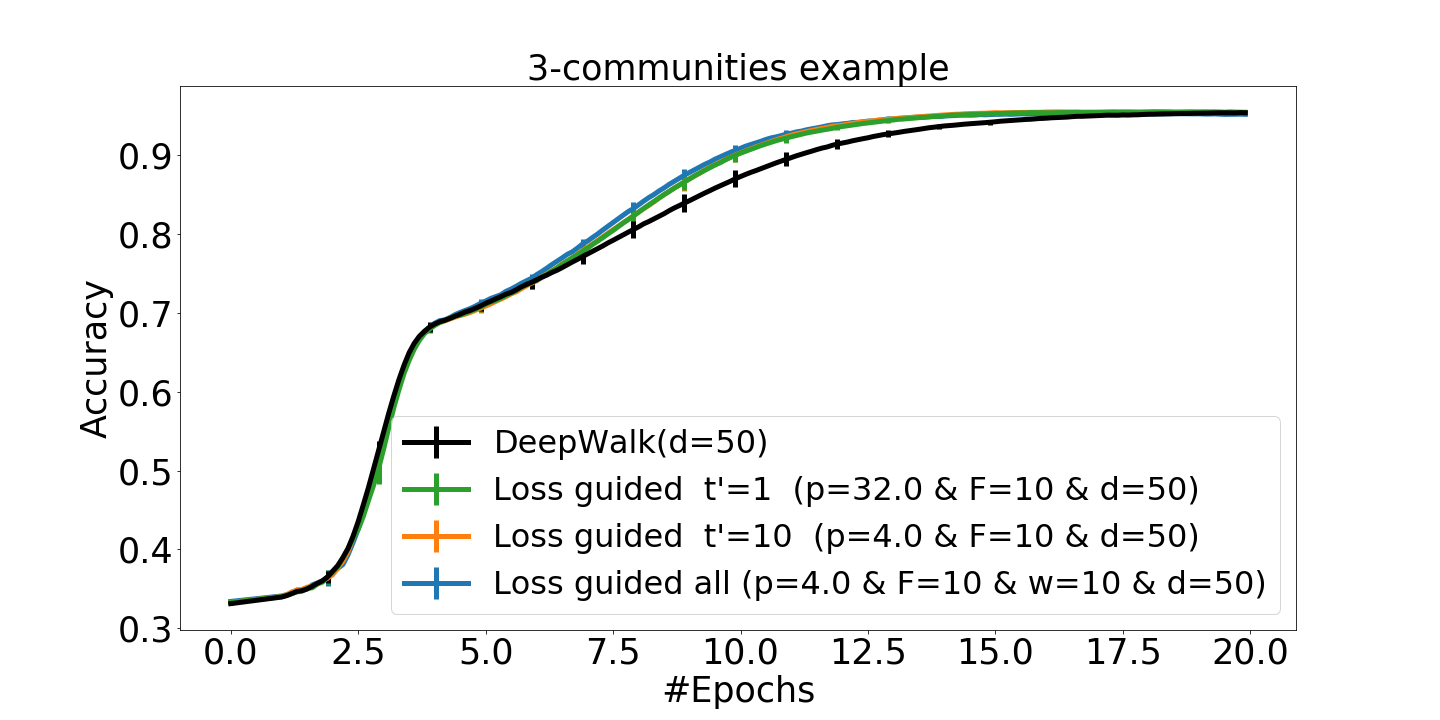}
    \caption{Example network: Accuracy in the course of training using {\sc DeepWalk} and loss-guided selection with {\sc DeepWalk} baseline. $d=10$ (left) and $d=50$ (right). The $x$-axis shows the number of epochs.}
    
    \label{plot:example_train}
\end{figure}

\begin{figure}
    \includegraphics[width=0.37\columnwidth]{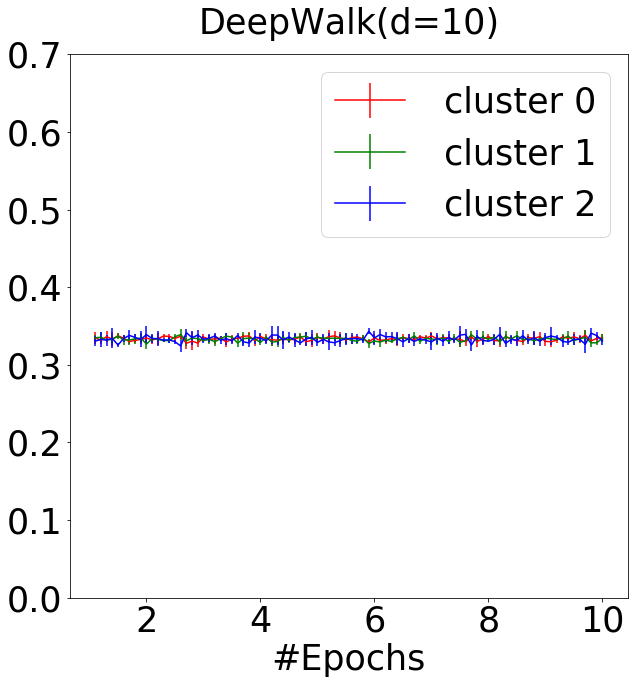}
    \includegraphics[width=0.45\columnwidth]{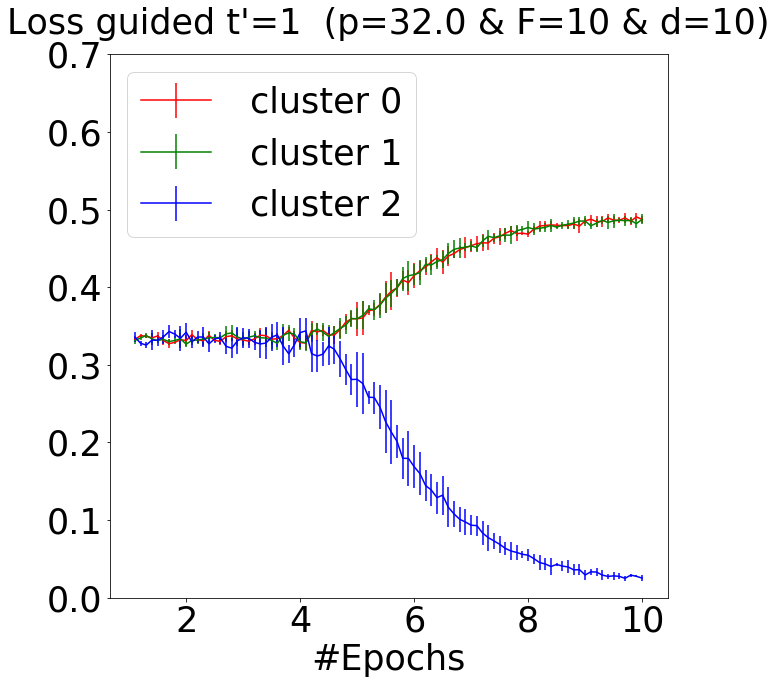}\\
\includegraphics[width=0.37\columnwidth]{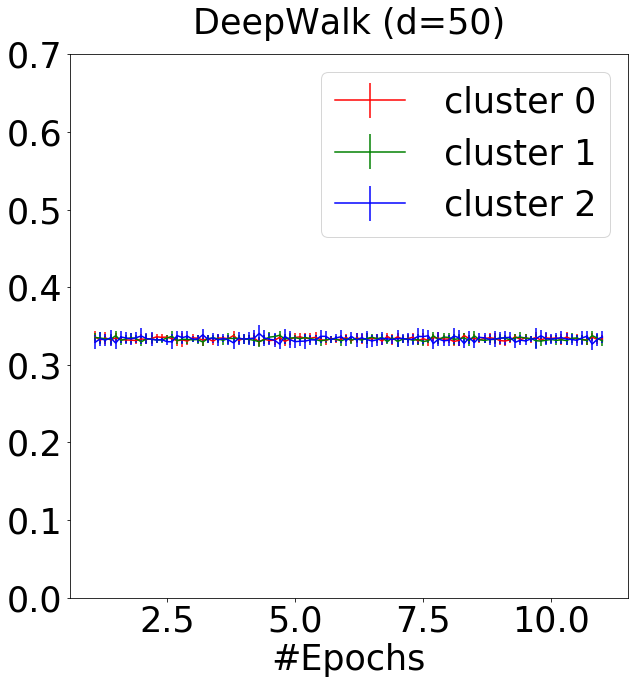}
    \includegraphics[width=0.45\columnwidth]{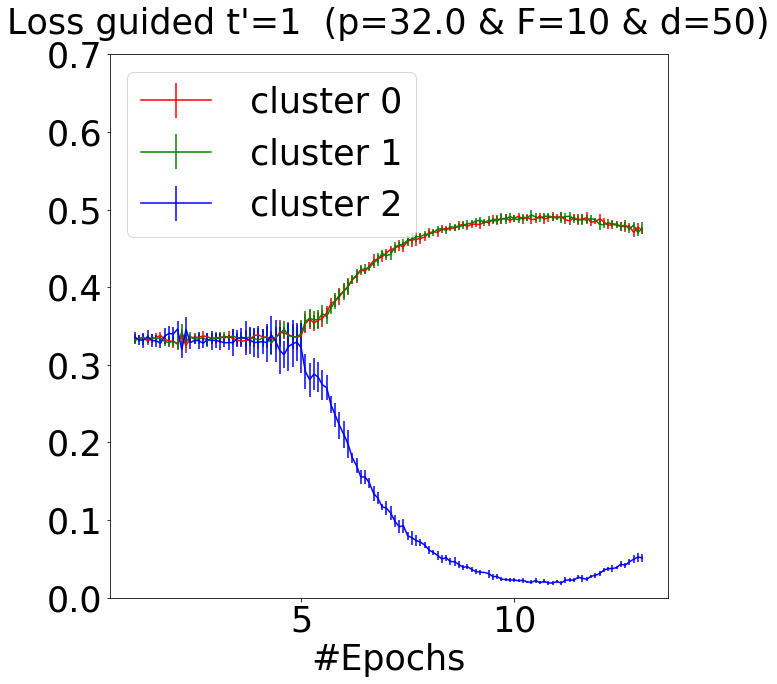}
        \caption{Example network: Fraction of training spent on walks starting in each community, in the course of training. {\sc DeepWalk} (left) and loss-guided (right). $d=10$ (top) and $d=50$ (bottom).}
    \label{plot:example_train_split}
\end{figure}

We trained node embeddings using {\sc DeepWalk} and using loss-guided selection with {\sc DeepWalk} as a baseline. 
The baseline method {\sc DeepWalk} selects a start node of a random walk uniformly and hence the distribution of training examples remains balanced among the three communities through the course of training.  The loss-guided selection will focus more training on walks with a higher loss score.  We expect the isolated community to separate out early in training and for the two inter-connected communities to require more training to "separate" from each other. The loss $L_+(i,j)$ of a same-community pair $(i,j)$ will be lower earlier for the isolated community.   A loss-guided method after the initial stage of training is more likely to select training examples from the two inter-connected communities and thus be more productive.  The benefit is further boosted by the corresponding selection of negative examples, where a community not selected for positive examples also does not participate in negative examples.   
The quality was measured by treating the problem as a 3-class classification problem as explained in Section~\ref{sec:metrics} with classes assigned according to community with $L=\{\text{red}, \text{green}, \text{blue}\}$.  Half the nodes (selected randomly) were used as labeled examples for the supervised component.
We used $r=25$ repetitions for each method and report representative results for embedding dimension $d=10,50$.
Figure~\ref{plot:example_train} shows the fraction of correct classifications as a function of training epochs.  We observe that the different methods behave the same in the initial phase of training, until the blue community separates out from the other two but after that the loss-guided methods are more effective.  
The loss-score function 
$\Lscore_1$ that uses the first edge of the walk attains the full advantage of the loss guided methods. This because the first edge identifies the community.
Figure~\ref{plot:example_train_split} reports the fraction of training spent at each community. We can see that in the initial phase all methods are balanced but as expected, the baseline {\sc DeepWalk} remains balanced whereas the loss-guided variant spend increasing fraction of training resources on the green and red communities, where it is more helpful.

\section{Hyperparameter sensitivity} \label{param_sensitivity:sec}

We explore the dependence of the performance of our loss-guided methods on the following parameters:   The number of edges $t'$ out of the $t=10$ walk edges that are used in the walk loss scoring function $\Lscore_{t'}$, the number $F$ of rounds per  epoch,  and the loss power value $p$ which determines how we weigh the loss of examples when we compute loss scores of walks.

\subsection{Loss scoring methods of walks} \label{comparing_approx_to_exact:sec}
We proposed (see Section~\ref{lossscoring:sec}) several loss scoring functions of walks: $\Lscore_{t'}(S)$ for $t'\in [10]$ which uses the average loss of the first $t'$ edges of the walk $S$ and $\Lscore_{\text{all}}(S)$ which uses the average loss of all positive training examples generated from the walk $S$.   We observed empirically that $\Lscore_{\text{all}}$ rarely outperformed $\Lscore_{10}$, even in terms of training cost.   We note that 
due to technical reasons we used the expected loss on the selected walk $S$ (under random draws of $\Delta_i$) instead of the precise evaluation on the pairs generated from the selected walk. This could have impacted adversely the reported performance  of $\Lscore_{\text{all}}$.  

We explore the training and computation cost with $\Lscore_{t'}$ as we vary $t'\in\{1,3,5,10\}$.
Representative results (with $F=10$ and $p=32$) are reported in Table~\ref{tab:different_t} (We report results for datasets for which the error bars are small compared with the gain and its variation.)  We see a general trend of improved training cost as we increase $t'$, but the extent of this improvement widely varies between datasets.   For example, the improvement is small for the TV shows dataset, moderate for the Politicians dataset, and significant for the Government and PPI datasets.  Note that the per-epoch computation cost also increases with $t'$ (see analysis in Section~\ref{complexity:sec} and Section~\ref{sec:gain}).  The overall computation gain as we increase $t'$ reflects both the decrease in the number of epochs and the increase in per-epoch computation.  We can see that the computation gain is often maximized for lower values of $t'$ than the value that maximizes the training gain.

\begin{table}[]
    \centering
    \scriptsize
    \begin{tabular}{c|c|c c | c|c|c}
      &  & & & \multicolumn{2}{c}{Training } & Comput  \\
        dataset & $t'$ & $p$ & $F$ &  \%gain & \%SD & \%gain  \\
        \hline
        \multicolumn{7}{c}{{\sc DeepWalk} baseline}\\
        \hline
PPI&$1$ & $32$ & $10$&12.7&3.91&10.4 \\ 
 &$3$ &  &   &20.7&3.77& \textbf{14.8} \\
 &$5$ &   &  &23.3&3.90&14.0 \\
 &$10$ &  &  & \textbf{25.9} &3.20&8.50 \\
\hline
\hline
Pubmed&$1$ & $32$ & $10$&\textbf{9.07}&3.91&\textbf{7.38}\\ 
&$2$ &  & &7.31&5.77&4.11\\ 
&$5$ &  & &1.49&9.00&-6.83\\
&$10$ &  & &-10.5&19.8&-29.9\\
\hline
\hline
Athletes&$1$ & $32$ & $10$&12.0&1.89&9.82\\ 
&$2$ &  & &13.5&2.34&\textbf{9.54}\\ 
&$5$ &  & &\textbf{16.8}&2.60&7.66\\ 
&$10$ &  & &15.5&4.49&-2.62\\
\hline
Company&$1$ & $32$ & $10$&18.2&1.86&\textbf{16.0}\\ 
&$2$ &  & &18.4&2.06&14.5\\ 
&$5$ &  & &\textbf{20.6}&1.47&11.8\\ 
&$10$ &  & &20.1&3.34&2.87\\
\hline
Government&$1$ & $32$ & $10$&10.7&2.67&8.47\\ 
&$2$ &  & &12.7&3.17&8.67\\ 
&$5$ &  & &19.7&2.99&\textbf{10.7}\\
&$10$ &  & &\textbf{23.6}&2.53&7.11\\ 
\hline
Politicians&$1$ & $32$ & $10$&17.9&2.19&\textbf{15.8}\\ 
&$2$ &  & &19.5&2.21&\textbf{15.8}\\ 
&$5$ &  & &23.6&1.75&15.6\\ 
&$10$ &  & &\textbf{24.8}&1.38&9.72\\
\hline
Public figures&$1$ & $32$ & $10$&10.2&5.08&7.93\\ 
&$2$ &  & &12.2&4.93&8.10\\ 
&$5$ &  & &23.0&1.84&\textbf{14.2}\\
&$10$ &  & &\textbf{26.9}&3.41&10.9\\
\hline
TV shows&$1$ & $32$ & $10$&21.8&1.14&\textbf{19.7}\\ 
&$2$ &  & &22.6&1.78&18.9\\ 
&$5$ &  & &23.8&1.43&15.7\\ 
&$10$ &  & &\textbf{24.4}&1.55&8.82\\

\hline
    \end{tabular}
\caption{Varying the loss scoring function $\Lscore_{t'}$ for $t'\in\{1,2,5,10\}$: Training gain and computation gain of selected datasets  ($F=10$, $p=32$, {\sc DeepWalk} baseline).}
\label{tab:different_t}
\end{table}

\subsection{Rounds per epoch $F$} 
The parameter $F$ controls the number of rounds per epoch. Recall that in each round we score $|V|$ walks and select $|V|/F$ of these walks for training. The setting $F=1$ corresponds to the baseline method.  In Table~\ref{tab:update_frequency} we report training and computation gains over the {\sc DeepWalk} baseline method for $F\in\{2,5,10,20\}$. We report on all datasets  for which the standard deviation on the gain allowed for meaningful comparisons.  Gains are reported with configurations $(\Lscore_1,p=32)$ and $(\Lscore_{10},p=4)$.
We highlight the value that maximizes the training cost or computation cost for each configuration.  We can see a trend where the training gain increases with  $F$.  We see that the computation gain is often maximized at a lower $F$ value than the value that maximizes the training gain. This is because the per-epoch computation also increases with $F$ (see Section~\ref{complexity:sec} and Section~\ref{sec:gain})~\eqref{compgain:eq}).  For some $F$ value we reach a sweet spot that balances the benefits from reduced training (that increase  with $F$) and the higher per-epoch pre-processing computation (that increases with $F$).
 
Qualitatively, higher $F$ values mean that the training is more focused on high loss examples and that the loss values are more current. 
This is helpful to some point, but with high enough $F$ we might direct all the training to outliers or erroneous examples.  In the table we do not see a point where the training cost starts increasing with $F$ but we do see that there is almost no gain between $F=10$ and $F=20$.

\begin{table}[]
    \centering
    \footnotesize
    \begin{tabular}{c|c|| c c|c || c c | c}
       &   & \multicolumn{2}{c}{Training } & Comp  & \multicolumn{2}{c}{Training } & Comp \\
        dataset& $F$ &  \%gain & \%SD & \%gain &  \%gain & \%SD & \%gain \\
        \hline
        & & \multicolumn{3}{c||}{$\Lscore_1$, $p=32$} & \multicolumn{3}{c}{$\Lscore_{10}$, $p=4$}   \\
        \hline
        
    PPI&  $2$ &10.4 & 5.30 &9.71 &16.7&5.20&\textbf{12.4}\\
      &   $5$& {\bf 13.2} &5.30& {\bf 11.8} &20.0&4.80&10.4 \\
      &   $10$&12.7&3.91&10.4 &\textbf{22.3}&3.38&4.06\\
      &   $20$&10.1&7.50&5.70 &21.5&3.40&-14.7 \\
    \hline
    \hline
    Athletes & $2$&7.04&4.30&6.46 &10.4&4.80&6.35\\
      &  $5$& 11.7&2.57& {\bf 10.5} &17.5&3.13&\textbf{8.45}\\
    &   $10$& {\bf 12.0} &1.89&9.82 &18.2&3.09&0.70\\
    &   $20$& 9.93&2.51&5.85 &\textbf{18.6}&2.11&-15.8\\
    \hline
    Company  & $2$&12.0&2.93&11.3 &13.2&2.68&9.14 \\
    &  $5$&\textbf{19.2}&2.20&\textbf{17.8} &20.0&2.11&\textbf{11.1}\\
    &  $10$&18.2&1.86&16.1 &\textbf{22.3}&1.71&5.54\\
    &  $20$&18.0&1.69&14.1 &22.2&1.26&-10.8\\
    \hline
    Government & $2$&6.93&3.19&6.35 &15.3&4.06&11.3 \\
    &  $5$&10.5&3.31&\textbf{9.27} &20.3&2.47&\textbf{11.4} \\
    &  $10$&\textbf{10.7}&2.67&8.47 &22.0&1.89&5.06\\
    &  $20$&9.75&2.64&5.60 &\textbf{22.4}&1.86&-10.7\\
    \hline
     New Sites & $2$&13.3&2.95&12.6 &\textbf{15.6}&5.65&\textbf{11.4}\\
     & $5$&\textbf{17.1}&3.48&\textbf{15.7} &8.30&7.58&-2.40 \\
     & $10$&15.2&3.53&12.86 &4.21&9.58&-17.8 \\
     & $20$&11.1&4.60&6.81 &4.33&10.6&-39.5 \\
    \hline
    Politicians & $2$&11.2&4.55&10.6 &14.6&2.00&10.8 \\
    &  $5$&16.8&2.38&15.6 &21.9&2.14&\textbf{13.7}\\
    &  $10$&17.9&2.19&\textbf{15.8} &24.6&1.60&9.44 \\
    &  $20$&\textbf{19.1}&1.83&15.4 &\textbf{25.2}&0.88&-4.08\\
    \hline
    Public  & $2$&5.63&7.39&5.07 &15.5&4.58&11.5 \\
    Figures &  $5$&\textbf{10.8}&4.20&\textbf{9.56}&21.6&4.19&\textbf{12.7}  \\
    &  $10$&10.2&5.08&7.93 &24.3&2.97&7.69 \\
    &  $20$&7.66&3.12&3.41 &\textbf{26.3}&1.98&-5.30  \\
    \hline
    TV Shows & $2$&10.5&3.09&9.88 &17.4&3.59&13.6\\
    & $5$&20.7&1.40&\textbf{19.4} &23.3&1.72&\textbf{15.1}\\
    & $10$ & \textbf{21.8}&1.14&19.7 &25.2&1.51&9.90 \\
    & $20$&21.8&1.20&18.2 &\textbf{26.1}&1.34&-3.40 \\
\hline
    \end{tabular}
\caption{Varying the number of rounds per epoch $F\in\{2,5,10,20\}$: Training gain and computation gain of selected datasets with respect to {\sc DeepWalk} baseline.  For loss-guided with $(\Lscore_1,p=32)$ and $(\Lscore_{10},p=4)$.}
\label{tab:update_frequency}
\end{table}

 \begin{table}[t]
    \centering
    \footnotesize
    \begin{tabular}{c|c|| c c|c || c c | c}
       &   & \multicolumn{2}{c}{Training } & Comp  & \multicolumn{2}{c}{Training } & Comp \\
        dataset& $p$ &  \%gain & \%SD & \%gain &  \%gain & \%SD & \%gain \\
        \hline
        & & \multicolumn{3}{c||}{$\Lscore_1$} & \multicolumn{3}{c}{$\Lscore_{10}$} \\

        \hline

PPI& $1$ &6.02&4.33&3.71&10.9&3.80&-9.70\\
&  $4$ &\textbf{12.1}&4.10&\textbf{9.8}&22.3&3.38&4.06\\
&  $32$ & 12.7&3.91&10.4&\textbf{22.8}&3.00&\textbf{5.00}\\ 

\hline
\hline
Pubmed& $1$ & 7.31&4.16&5.62& 6.70&6.79&-8.70\\ 
&  $4$ & 8.80&4.92&7.11& \textbf{5.21}&7.02&\textbf{-10.5}\\
&  $32$ & \textbf{9.07}&3.91&\textbf{7.38}& -10.5&19.8&-29.9\\ 
\hline
\hline
Athletes& $1$ & 8.73&2.81&6.56& 12.6&2.88&-6.18\\ 
&  $4$ & \textbf{12.1}&1.89&\textbf{9.88}& \textbf{18.2}&3.09&\textbf{0.70}\\
&  $32$ & 12.0&1.89&9.88& 15.5&4.49&-2.62\\
 
\hline
Company&$1$ &16.8&2.43&14.6&17.9&1.75&0.13\\ 
& $4$ & \textbf{19.4}&1.83&\textbf{17.2}&\textbf{22.3}&1.71&\textbf{5.54}\\ 
& $32$ & 18.2&1.86&16.1&20.1&3.34&2.87\\ 
\hline
Government&$1$ &10.3&2.43&8.07&15.4&3.06&-2.98\\ 
&  $4$ & \textbf{11.7}&2.75&\textbf{9.45}&22.0&1.89&5.06\\ 
&  $32$ &10.7&2.67&8.47&\textbf{23.6}&2.53&\textbf{7.11}\\ 
\hline
New sites&$1$ &16.5&3.70&14.2&11.3&3.80&-8.70\\ 
& $4$ &\textbf{16.6}&3.70&\textbf{14.3}&\textbf{9.70}&5.60&\textbf{-10.7}\\ 
& $32$&15.2&3.50 &12.9&4.00&6.90&-17.7\\ 
\hline
Politicians&$1$ &15.7&2.52&13.6&18.5&1.60&1.98\\ 
&  $4$ &17.2&2.19&15.1&24.6&1.60&9.44\\ 
&  $32$ &\textbf{17.9}&2.19&\textbf{15.8}&\textbf{24.8}&1.40&\textbf{9.72}\\ 
\hline
Public &$1$ &\textbf{10.5}&3.48&\textbf{8.25}&17.5&3.21&-0.70\\ 
Figures & $4$ & 9.51&4.09&7.29&24.3&2.97&7.69\\
& $32$ & 10.2&5.08&7.93&\textbf{26.9}&3.41&\textbf{10.9}\\
\hline
TV shows&$1$ & 18.3&1.61&16.2&19.3&1.89&2.63\\ 
&  $4$ & 21.3&1.40&19.1&\textbf{25.2}&1.55&\textbf{9.88}\\ 
&  $32$ & \textbf{21.8}&1.14&\textbf{19.7}&24.4&1.55&8.82\\ 

\hline
    \end{tabular}
\caption{Varying the power $p\in \{1 ,4, 32\}$:
Training gain and computation gain of selected datasets with respect to  {\sc DeepWalk} baseline.  For loss-guided with $\Lscore_1$ and $\Lscore_{10}$. We keep the update frequency fixed, $F=10$.}
\label{tab:different_p}
\end{table}

\subsection{Loss power $p$}

The loss power $p$ is used in the walk loss scoring function (see Section~\ref{lossscoring:sec}).  The value of $p$
determines to what extent the training selection made in each round is biased towards examples with higher loss. 
A lower $p$ allows for broader selection of walks into a round and higher $p$ focuses the training more on highest loss score walks. In particular, $p=32$
means that we essentially select walks with highest-loss examples whereas $p=1$
means that we select walks for training proportionately to the loss values of their example(s).
 The loss power $p$ selection should be dependent on $F$, because lower $F$ allows for a broader selection of walks as well.  It also needs to be fitted to the $t'$ we use.
Table~\ref{tab:different_p} reports training and computation gains when we vary the loss power $p \in \{1, 4, 32\}$.  This for the loss scoring functions $\Lscore_1$ and $\Lscore_10$.  We can see that higher $p$ values, $4$ or $32$, perform better overall than $p=1$ and that the improvement are fairly robust to the particular choice of $p$.

\section{Results for {\sc Node2Vec} baseline} \label{moren2v:sec}

\begin{figure}[h]
\vskip 0.2in
\begin{center}
\includegraphics[width=0.48\columnwidth]{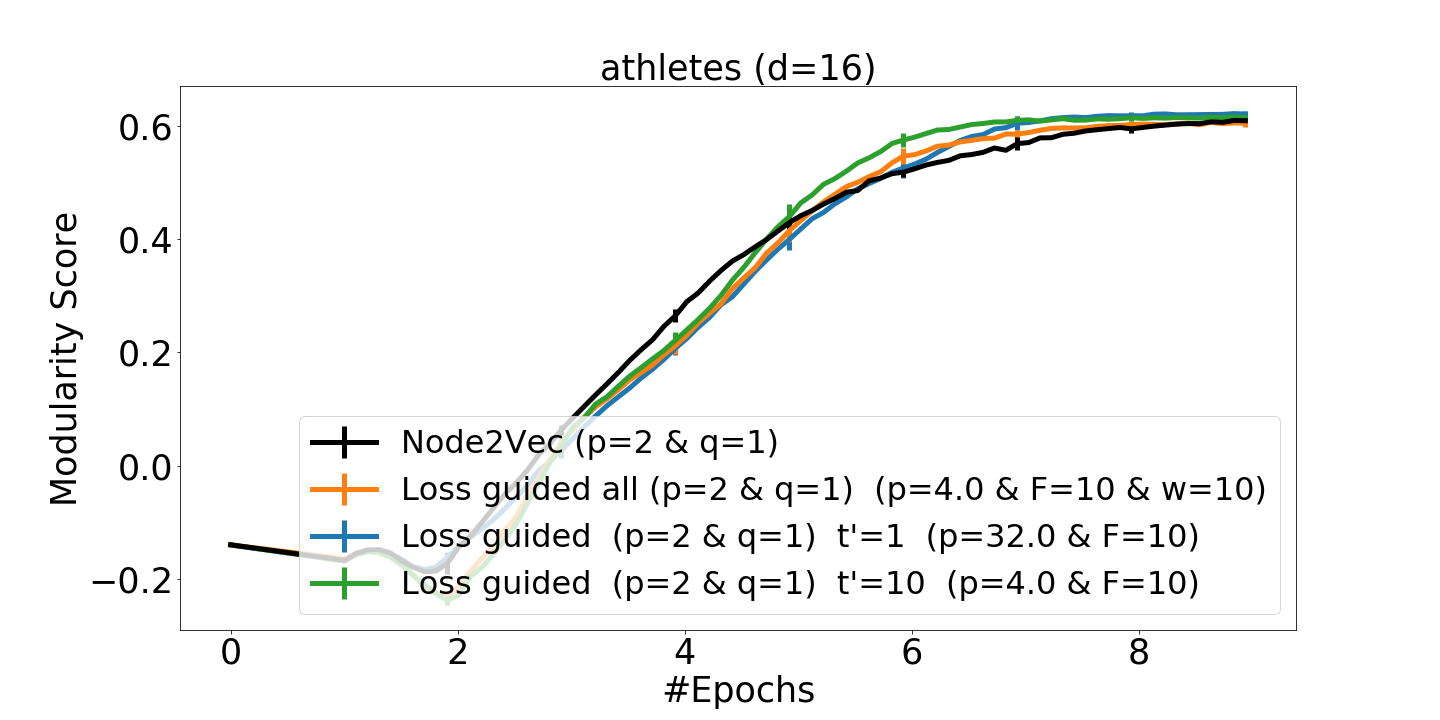}
 \includegraphics[width=0.48\columnwidth]{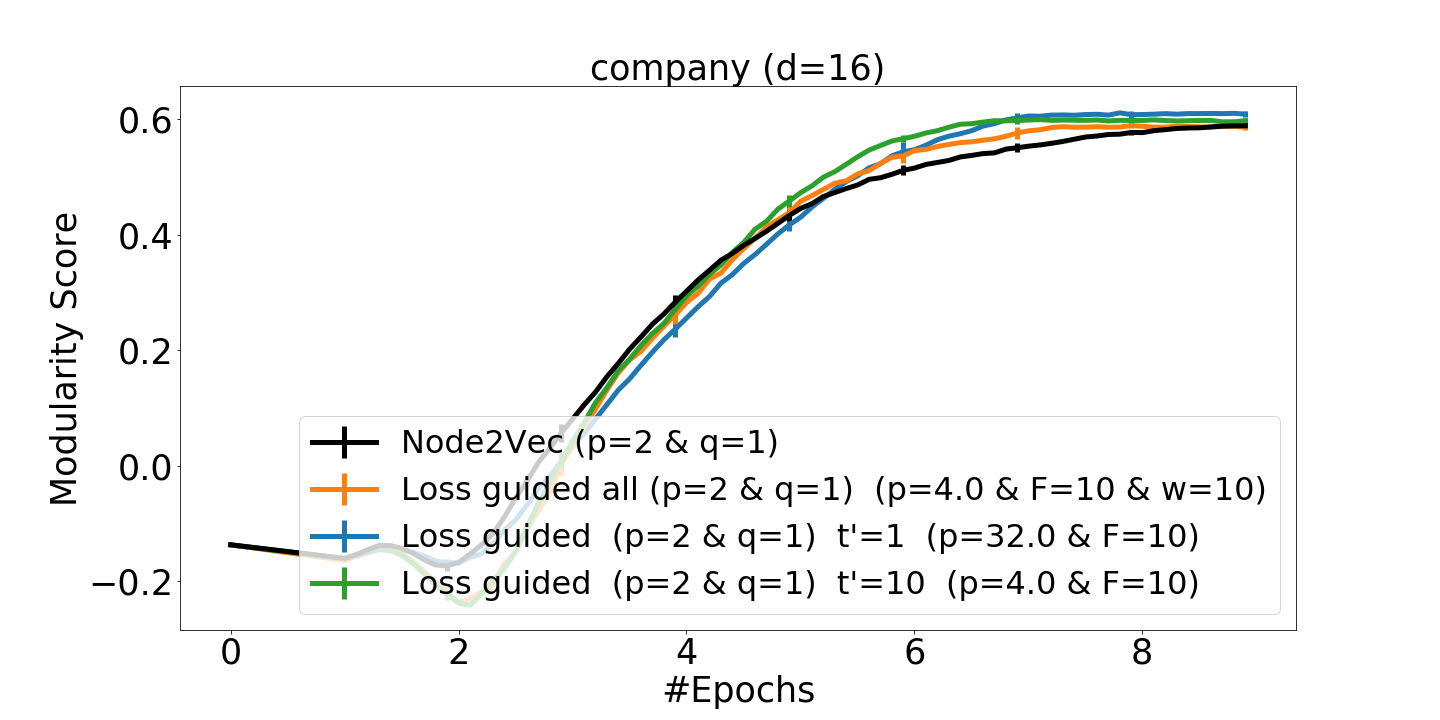}
\includegraphics[width=0.48\columnwidth]{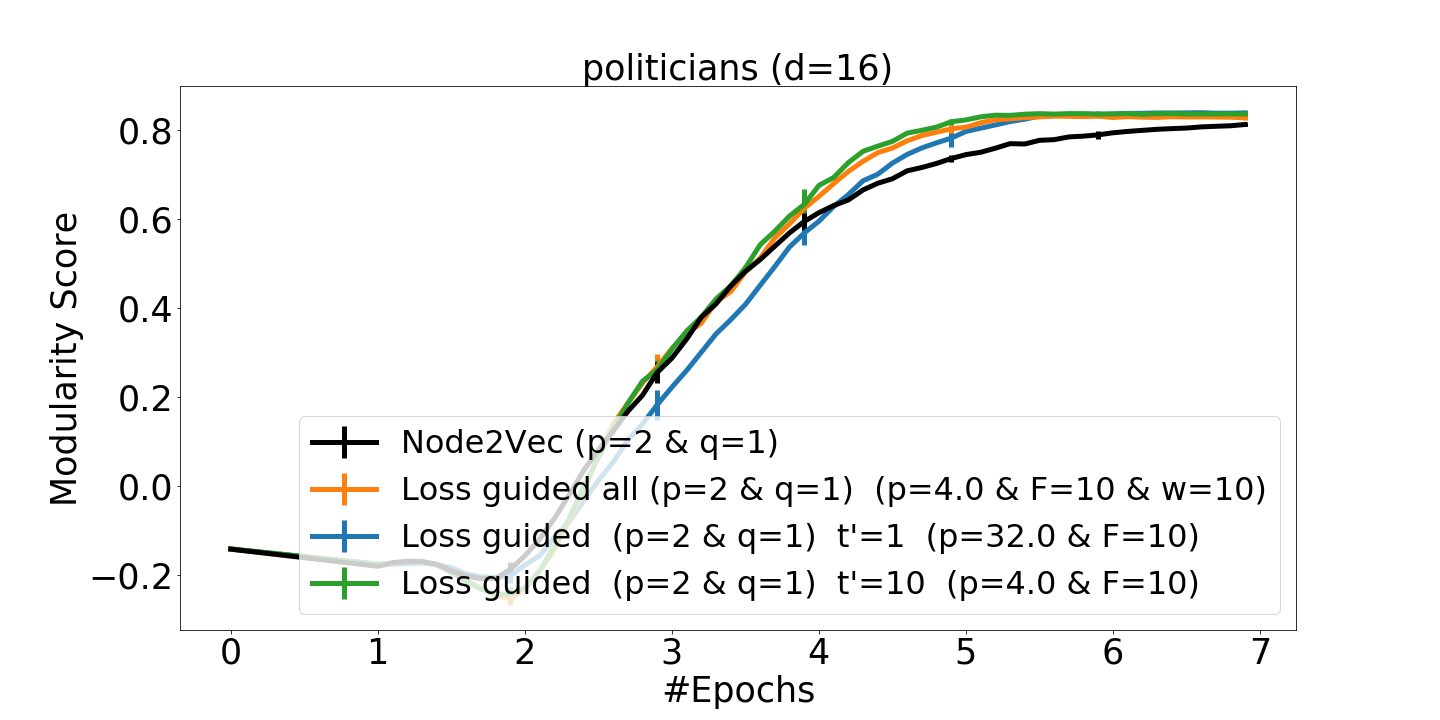}
\includegraphics[width=0.48\columnwidth]{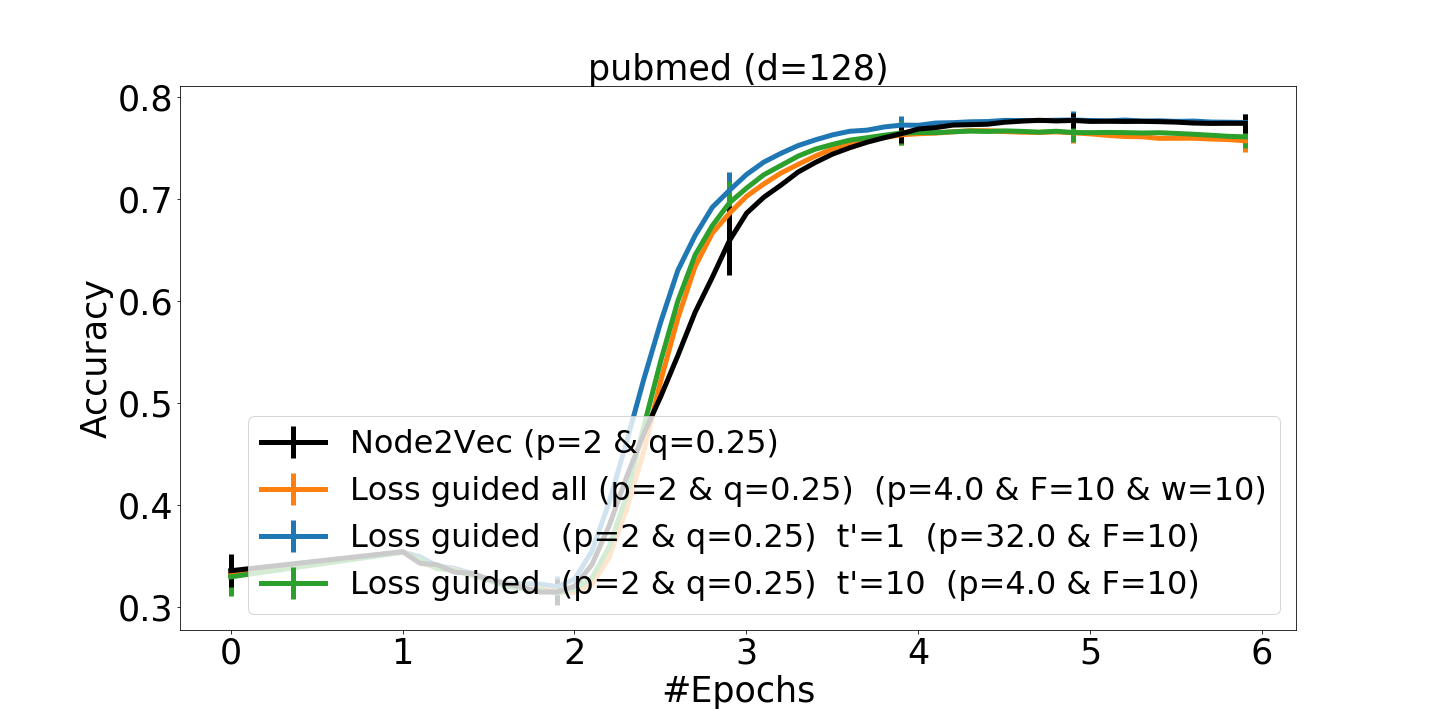}
\includegraphics[width=0.48\columnwidth]{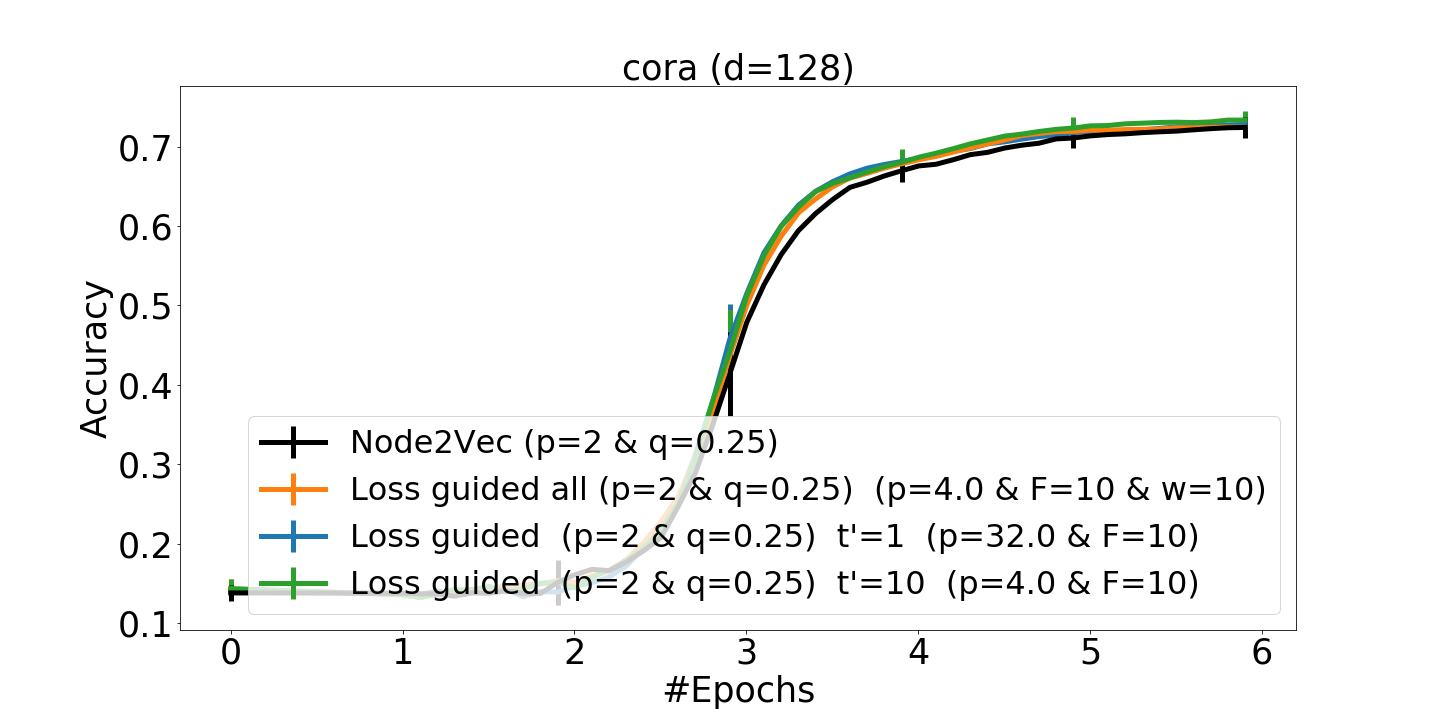}
\includegraphics[width=0.48\columnwidth]{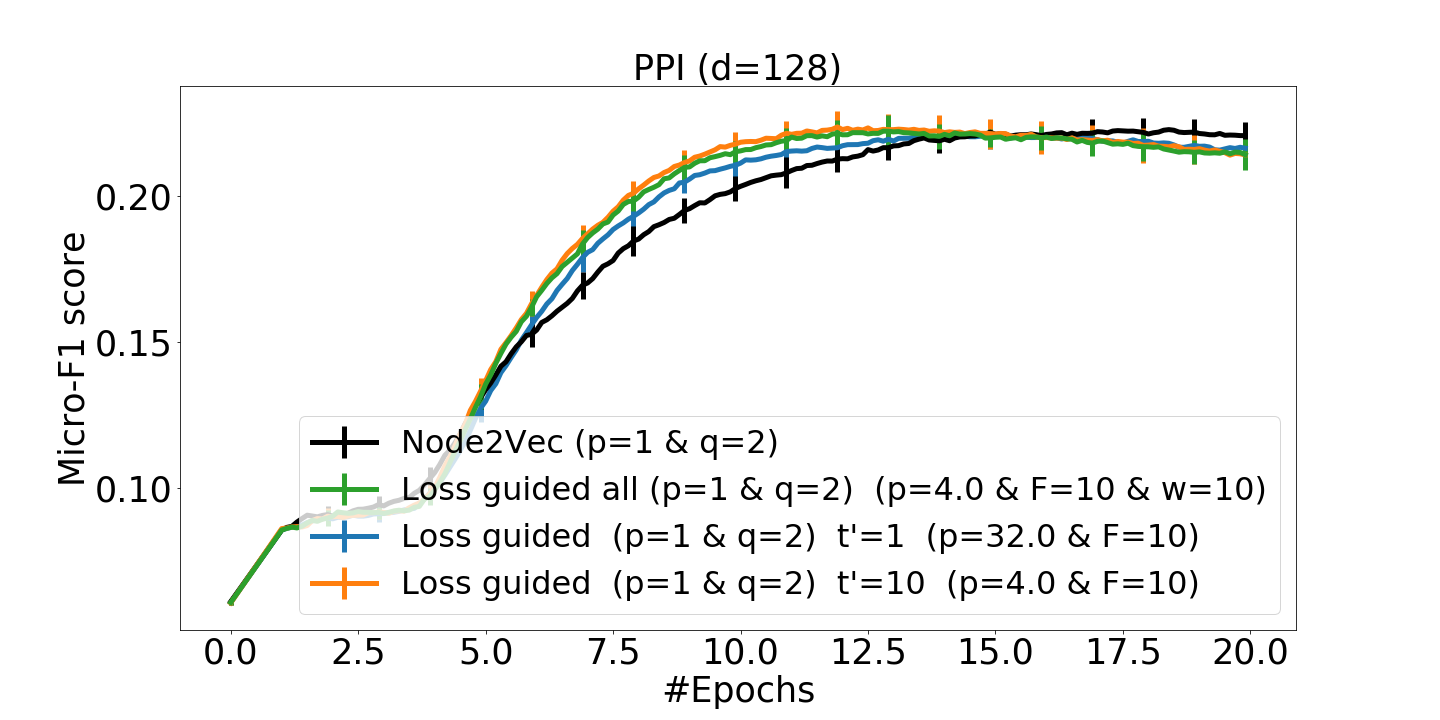}
\caption{{\sc Node2Vec} baseline: Quality in the course of training for the Athletes, Company, and Politicians networks from the Facebook collection, the Pubmed and Cora citation networks, and the PPI network.}
\label{plot:fb_n2v}
\end{center}
\vskip -0.2in
\end{figure}

Plots for the quality in the course of training with the {\sc Node2Vec} baseline for representative datasets are provided in Figure~\ref{plot:fb_n2v}.  Recall that the respective training and computation gains of the loss guided methods were reported in Table~\ref{tab:facebook_network_optimal_params_performance} and Table~\ref{tab:citation_network_optimal_params}.

\begin{figure*}[h]
\begin{center}
 \includegraphics[width=0.65\columnwidth]{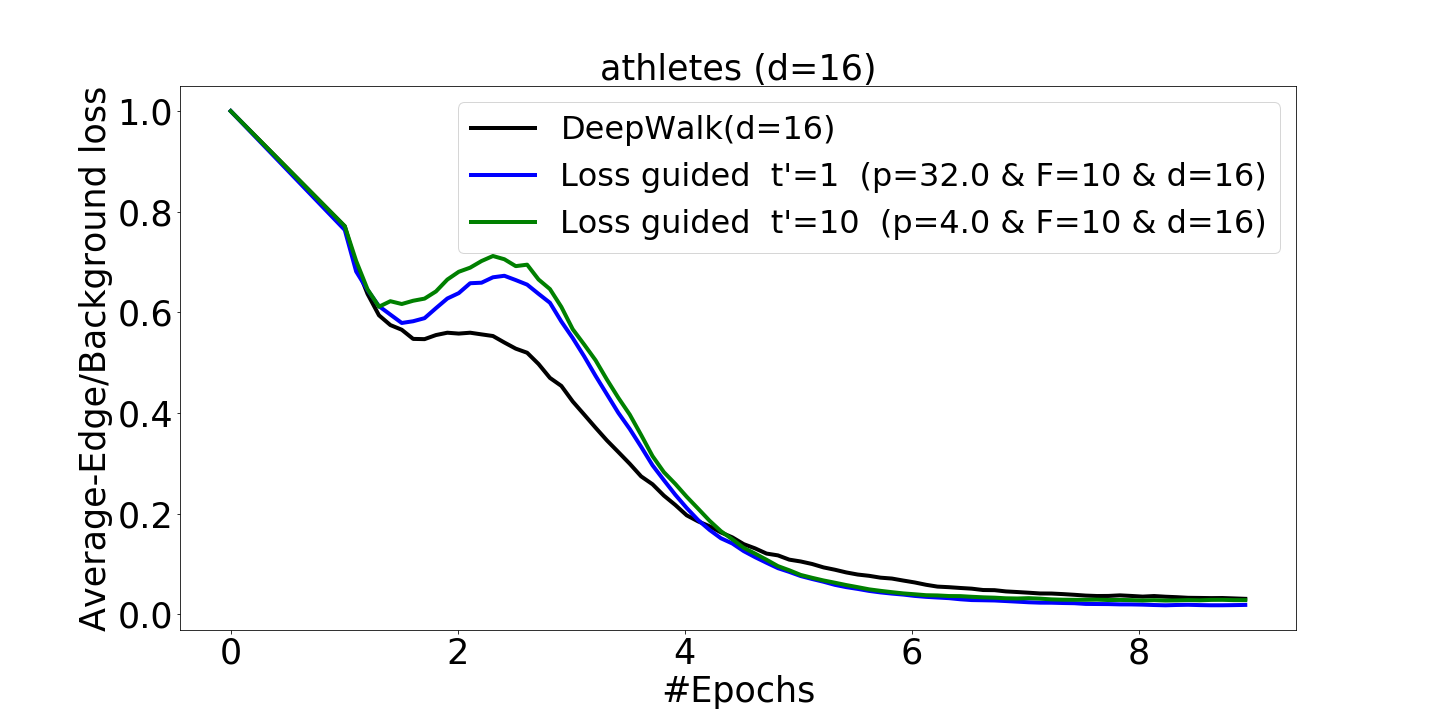}
 \includegraphics[width=0.65\columnwidth]{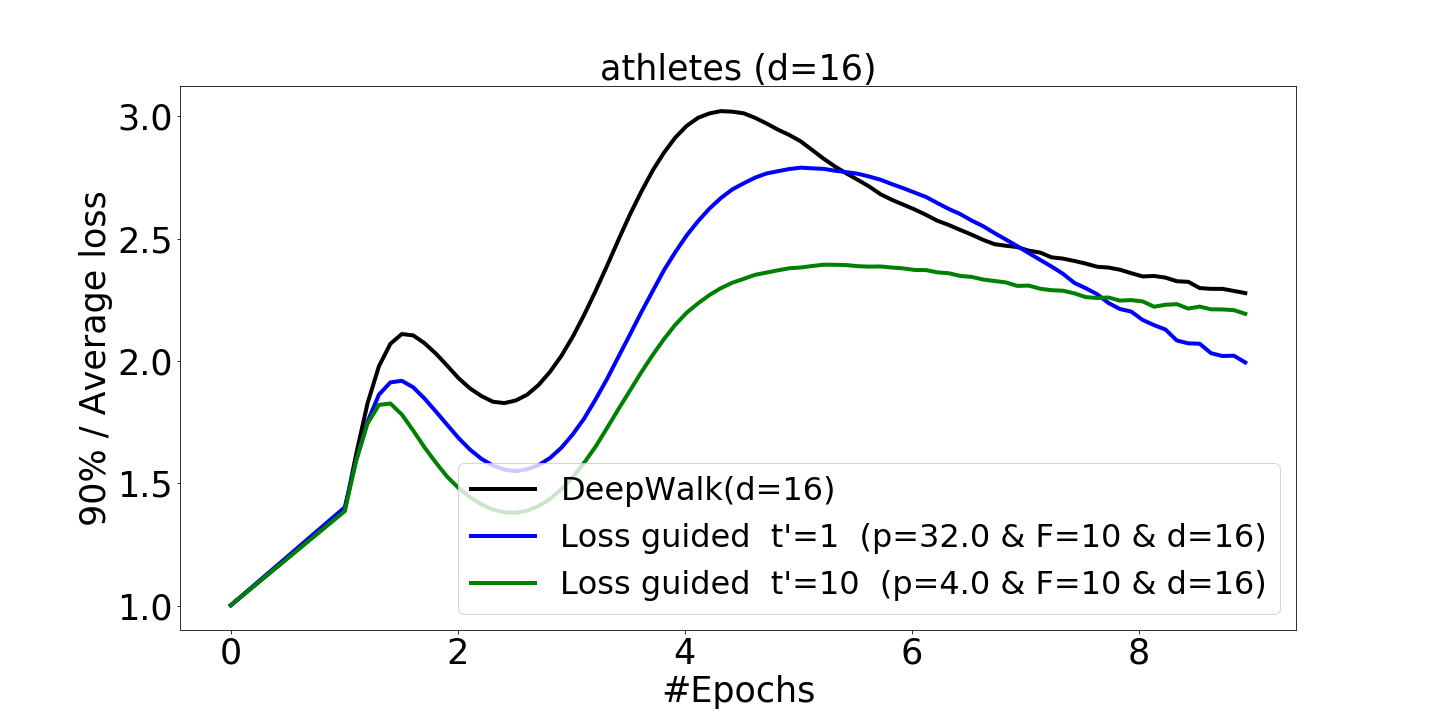}
 \includegraphics[width=0.65\columnwidth]{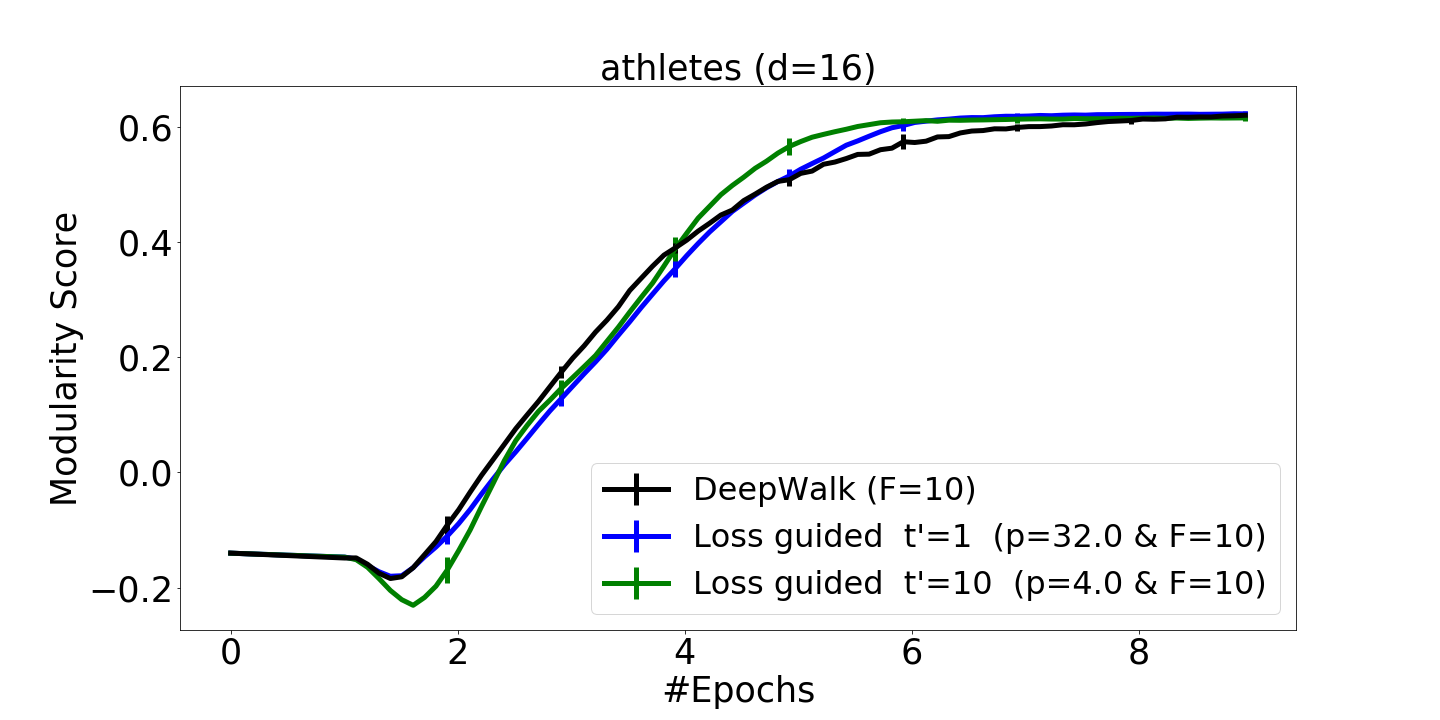}\\
  \includegraphics[width=0.65\columnwidth]{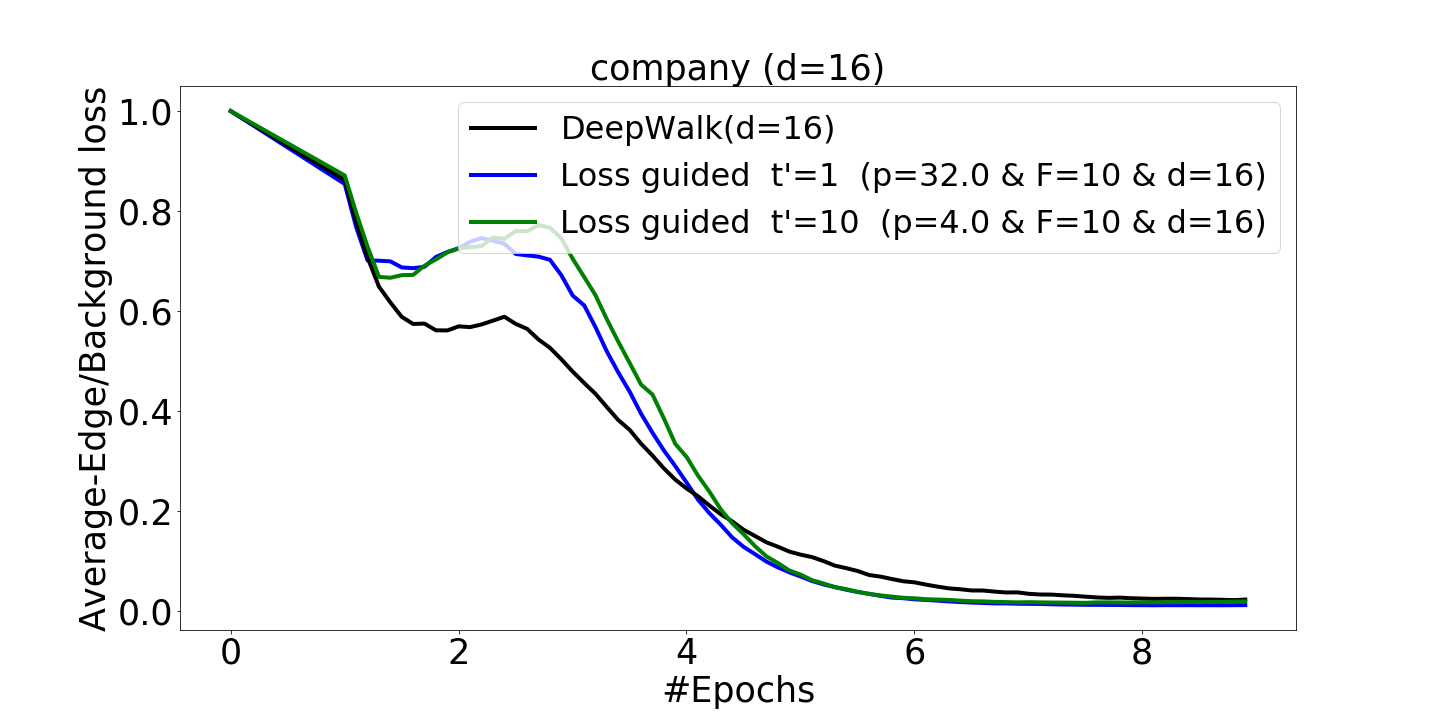}
 \includegraphics[width=0.65\columnwidth]{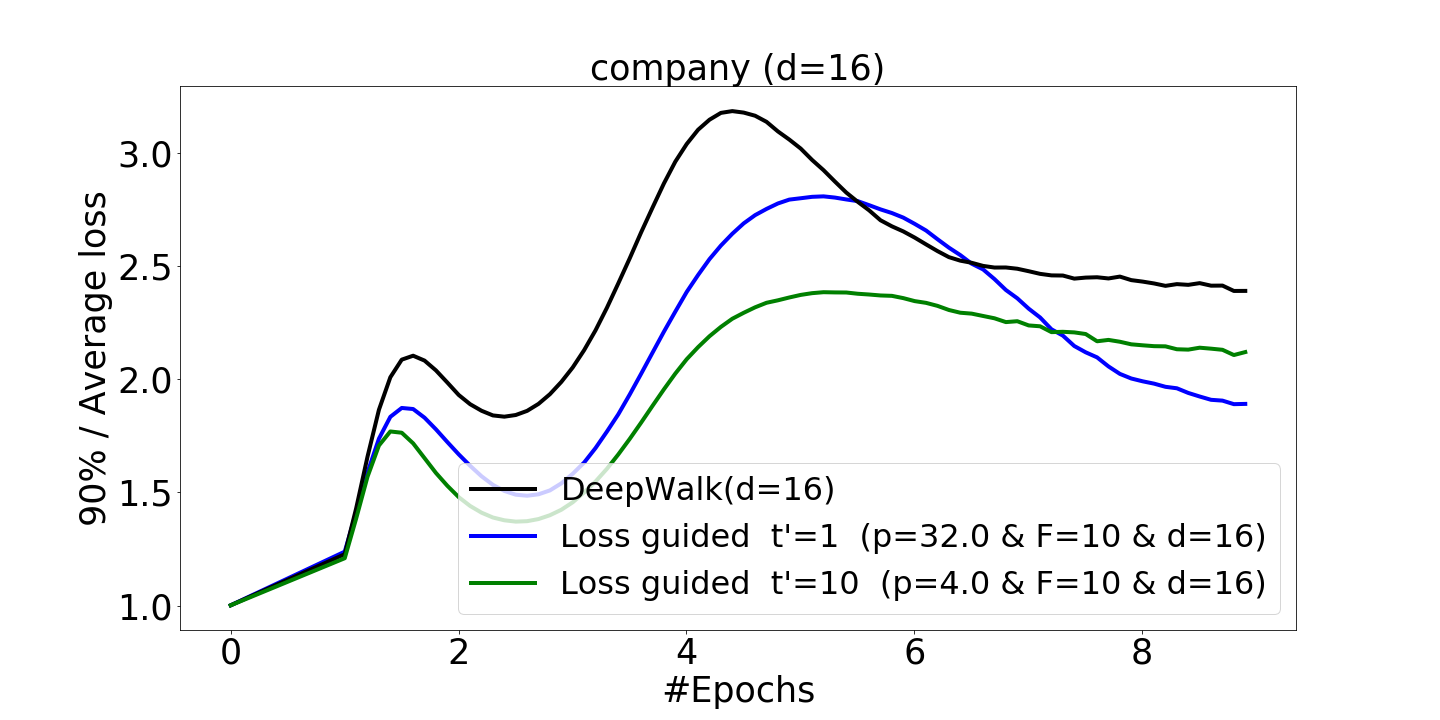}
 \includegraphics[width=0.65\columnwidth]{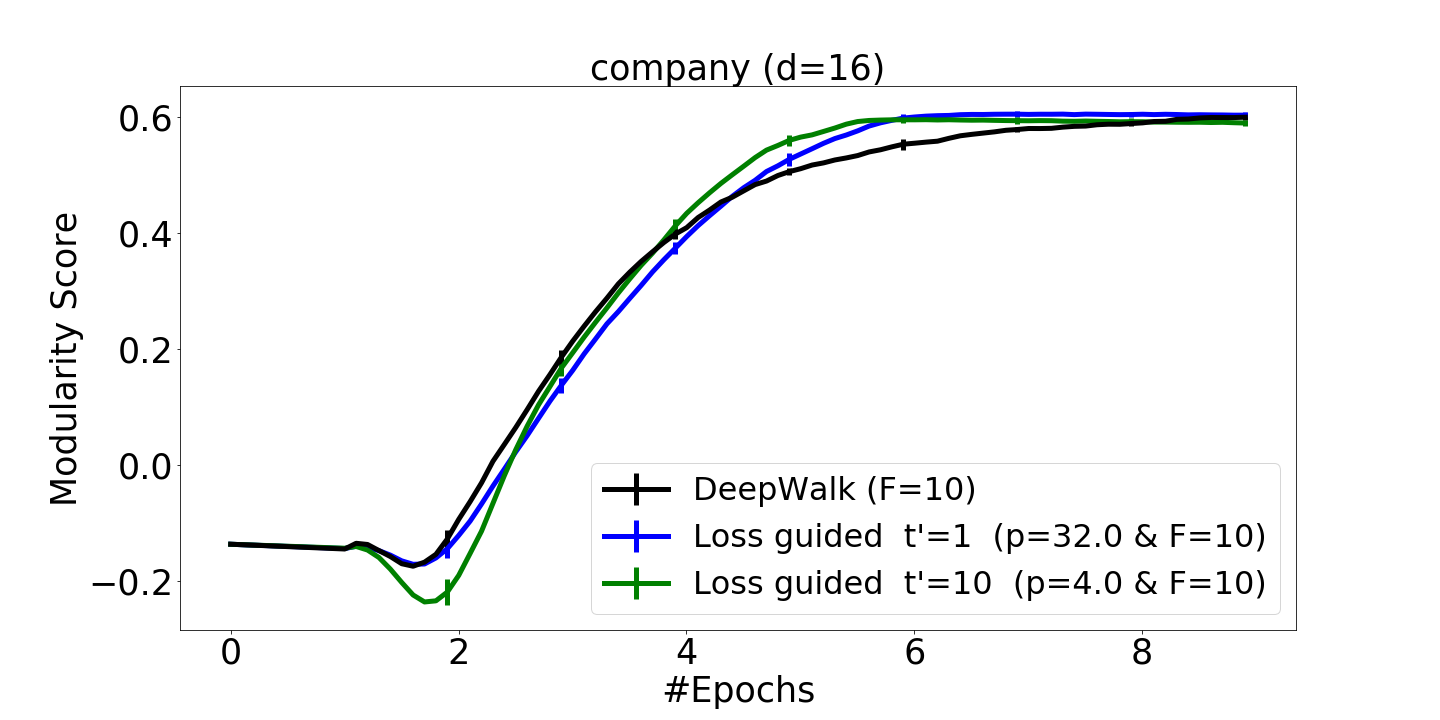}\\
 
 \includegraphics[width=0.65\columnwidth]{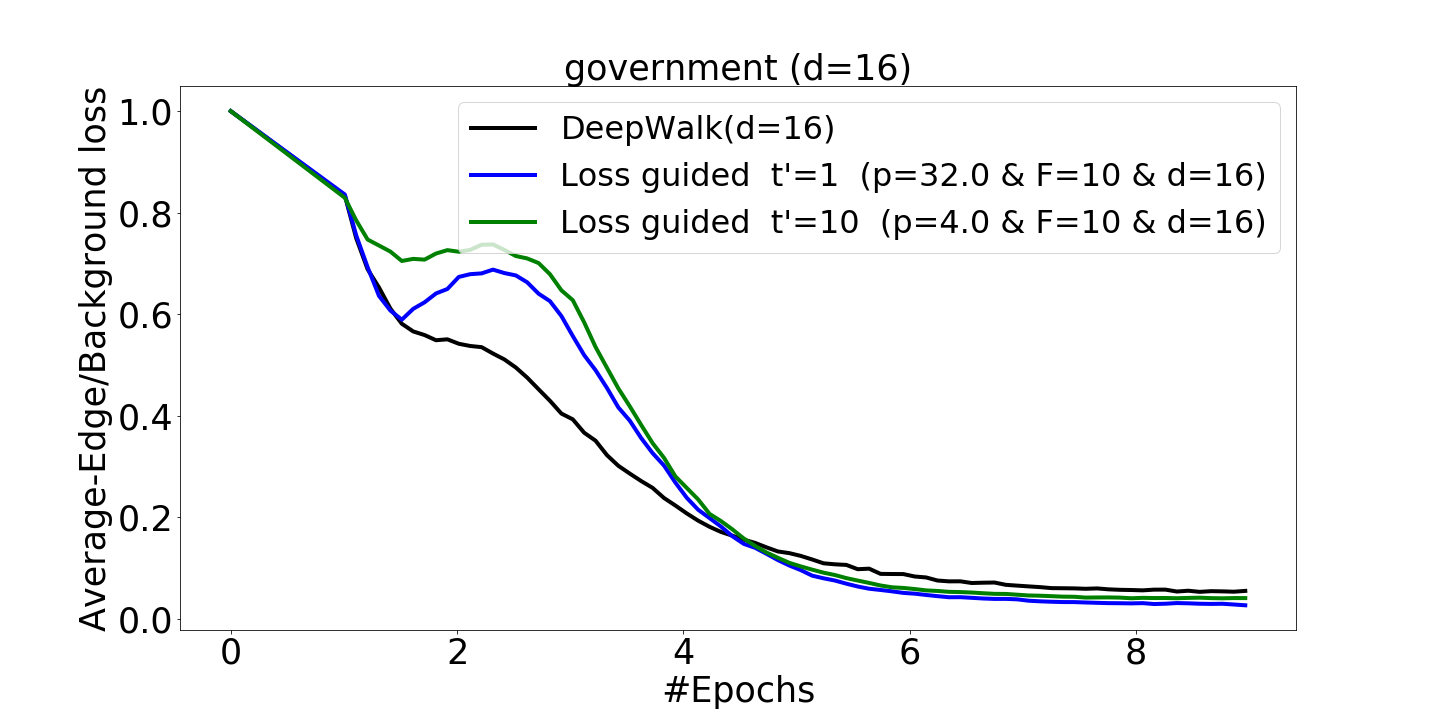}
 \includegraphics[width=0.65\columnwidth]{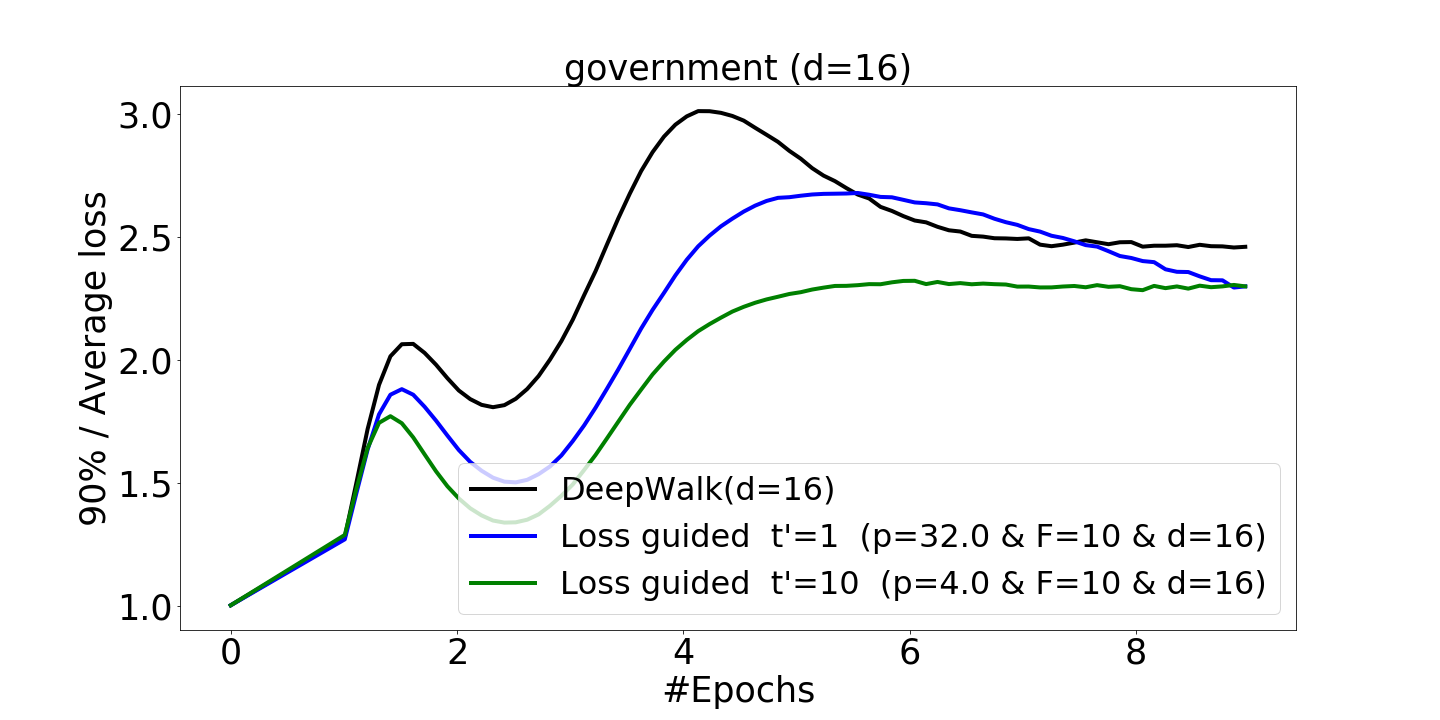}
 \includegraphics[width=0.65\columnwidth]{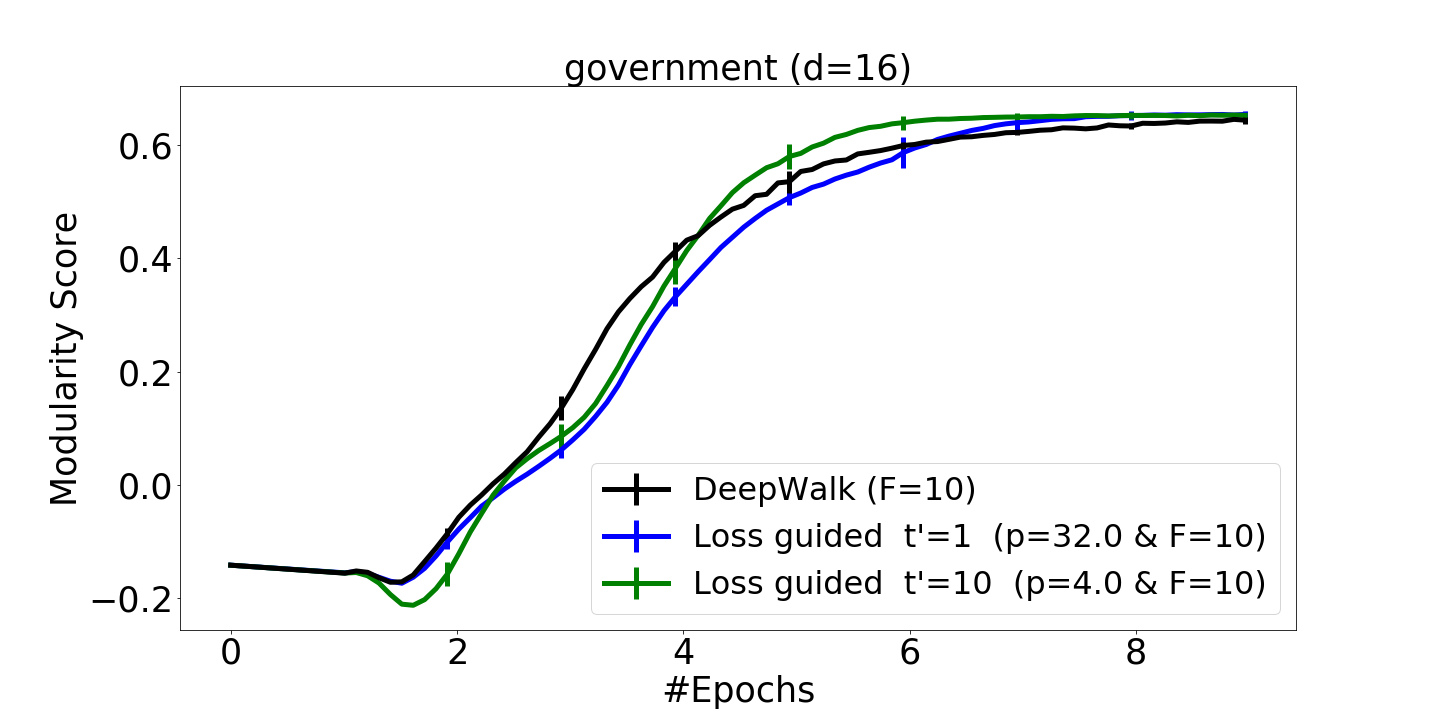}\\
 \notinproc{
 \includegraphics[width=0.65\columnwidth]{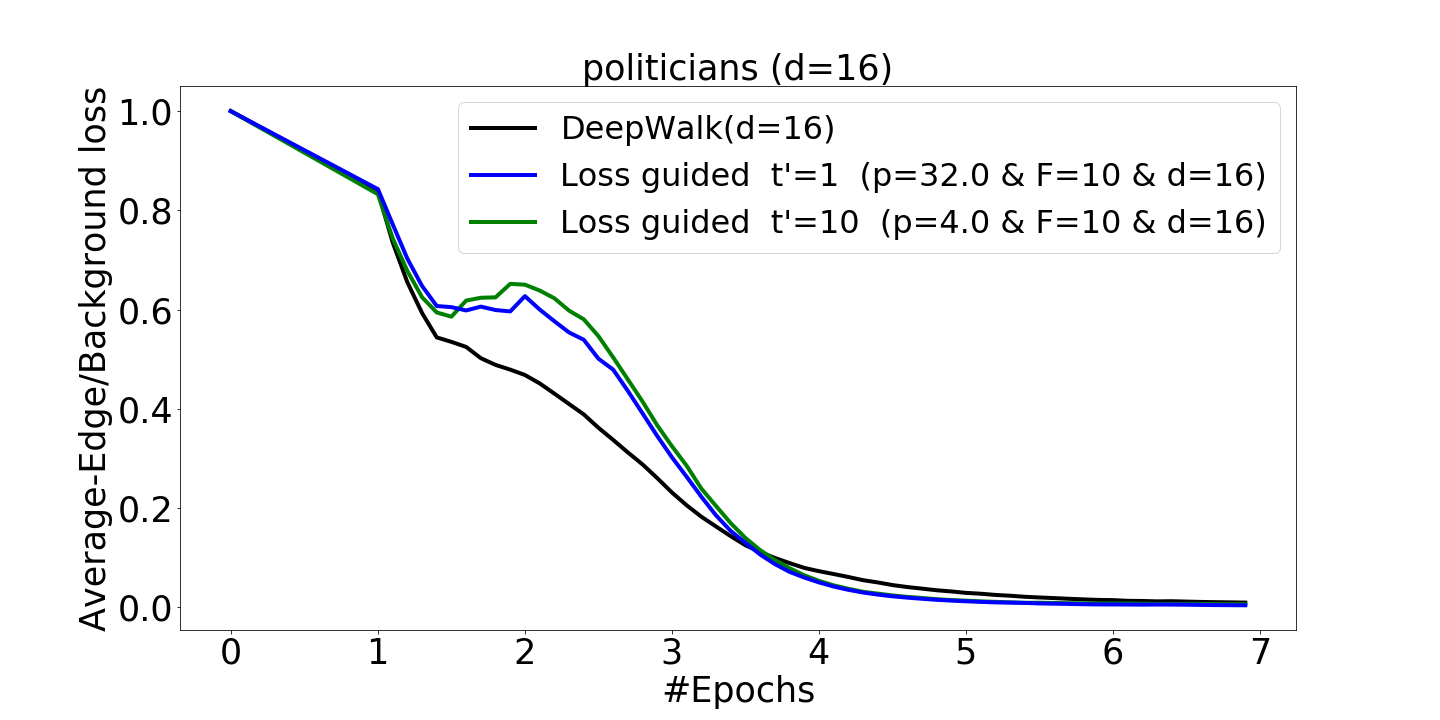}
 \includegraphics[width=0.65\columnwidth]{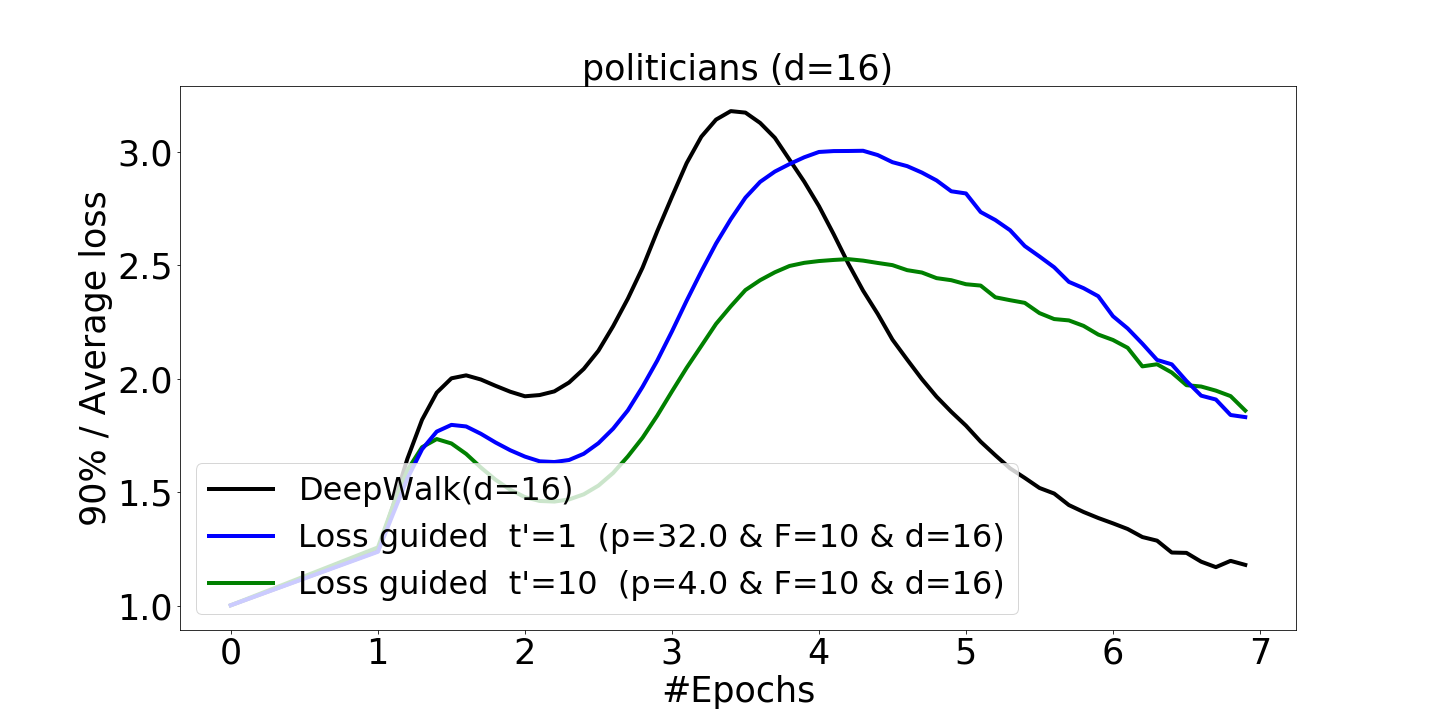}
 \includegraphics[width=0.65\columnwidth]{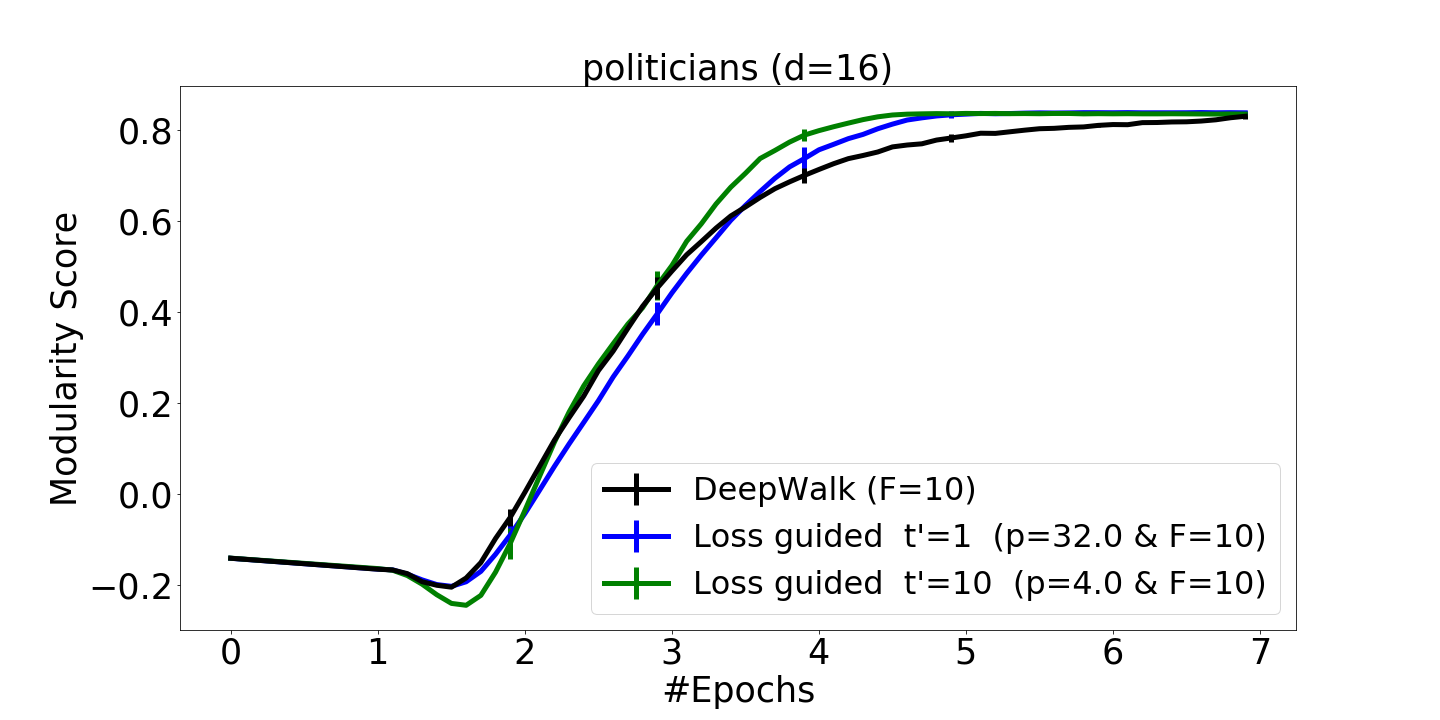}\\

 \includegraphics[width=0.65\columnwidth]{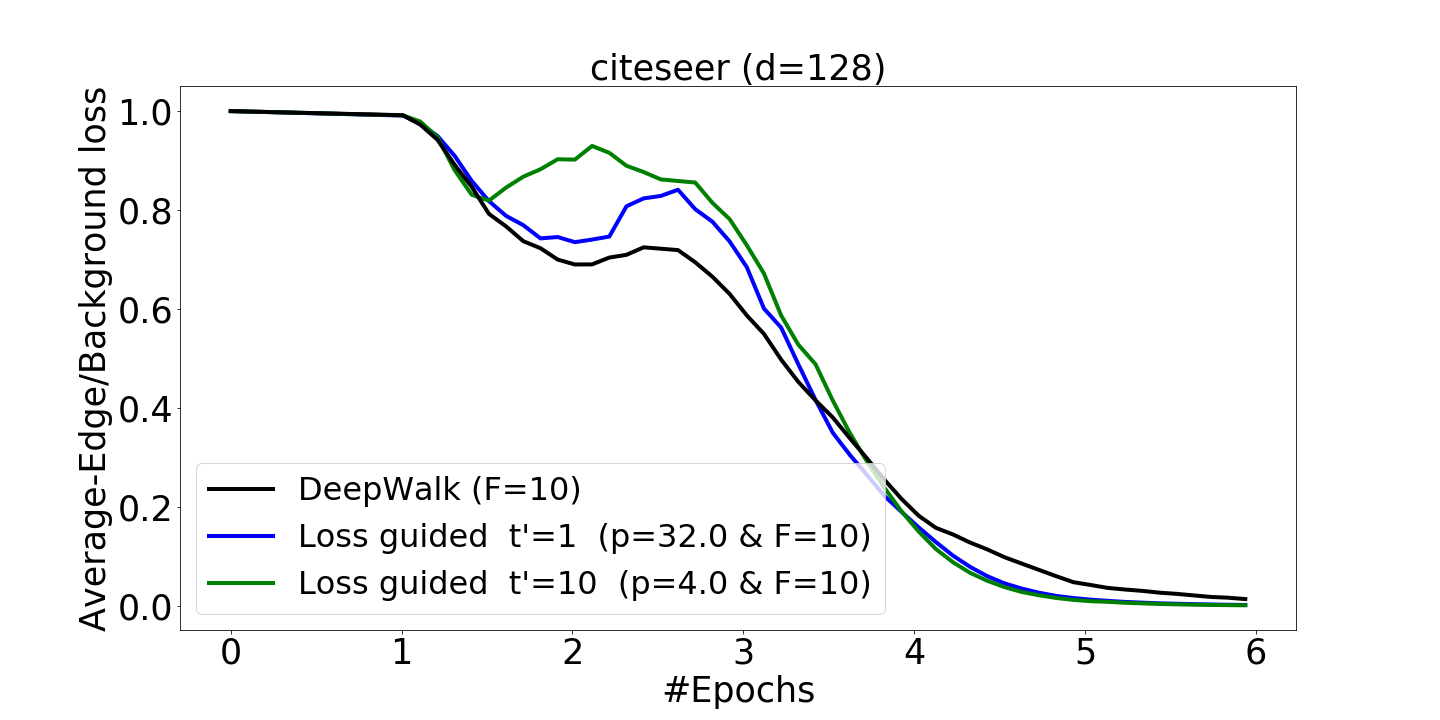}
 \includegraphics[width=0.65\columnwidth]{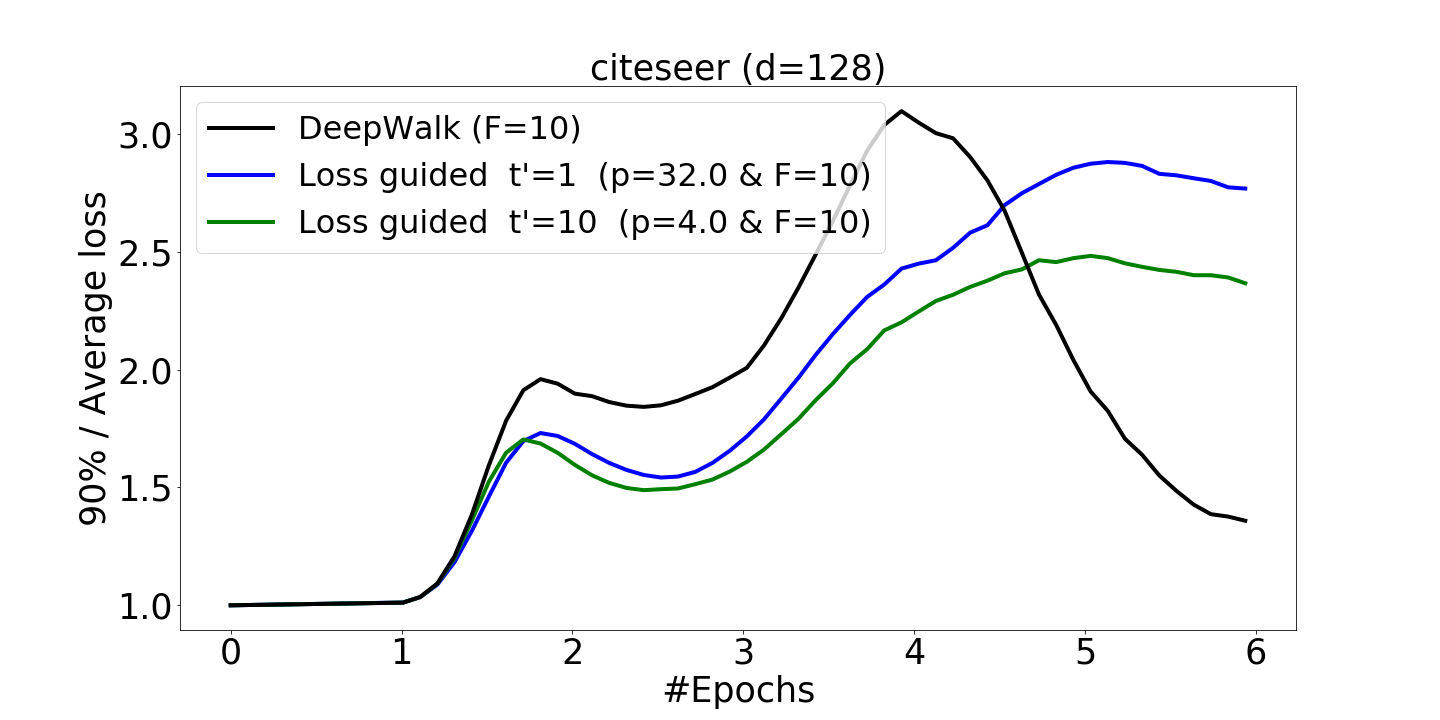}
 \includegraphics[width=0.65\columnwidth]{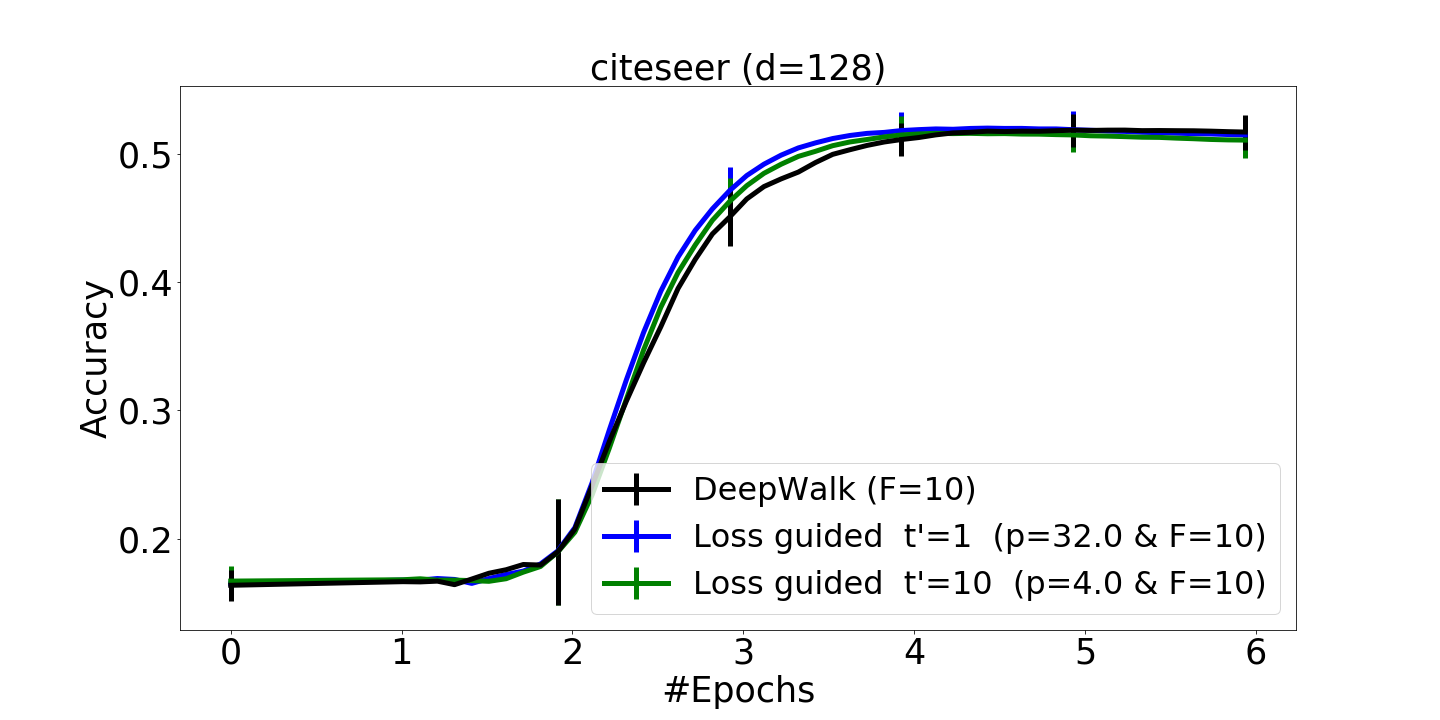}\\
 }

 \includegraphics[width=0.65\columnwidth]{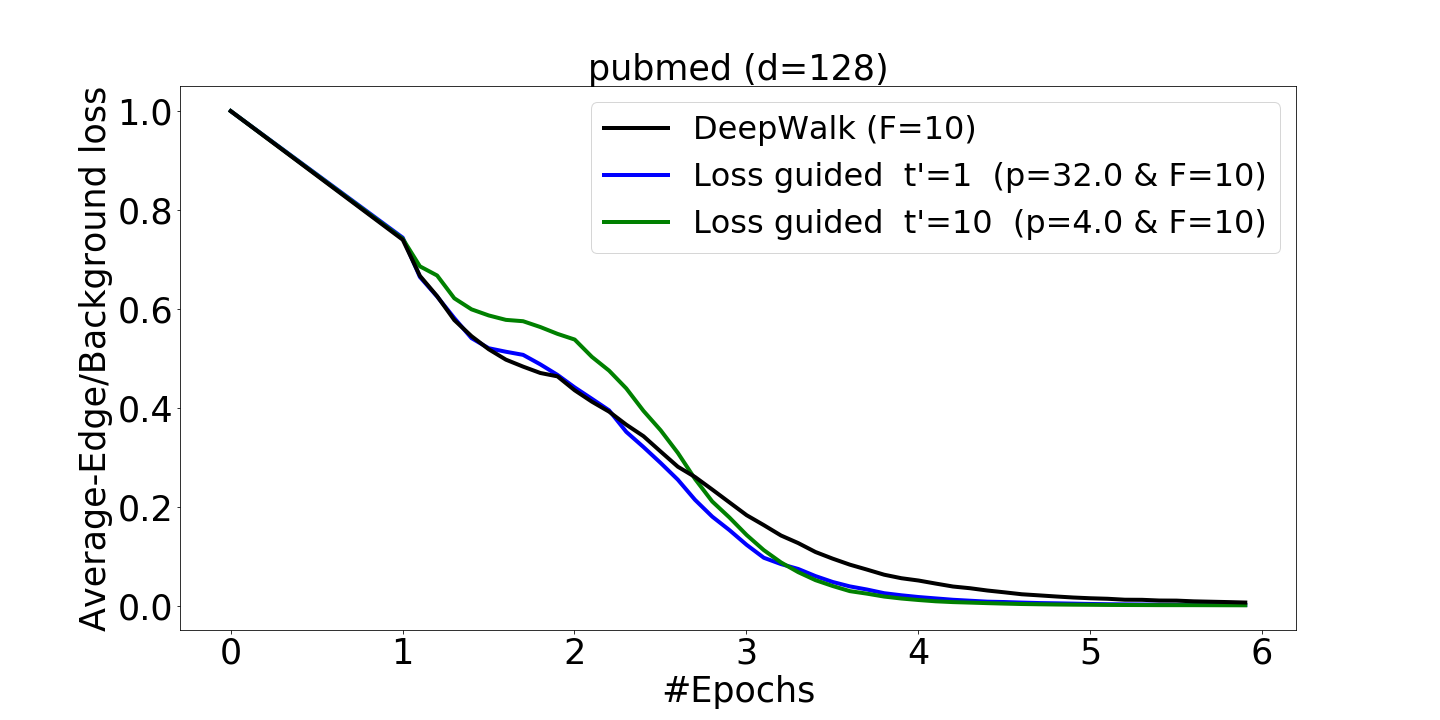}
 \includegraphics[width=0.65\columnwidth]{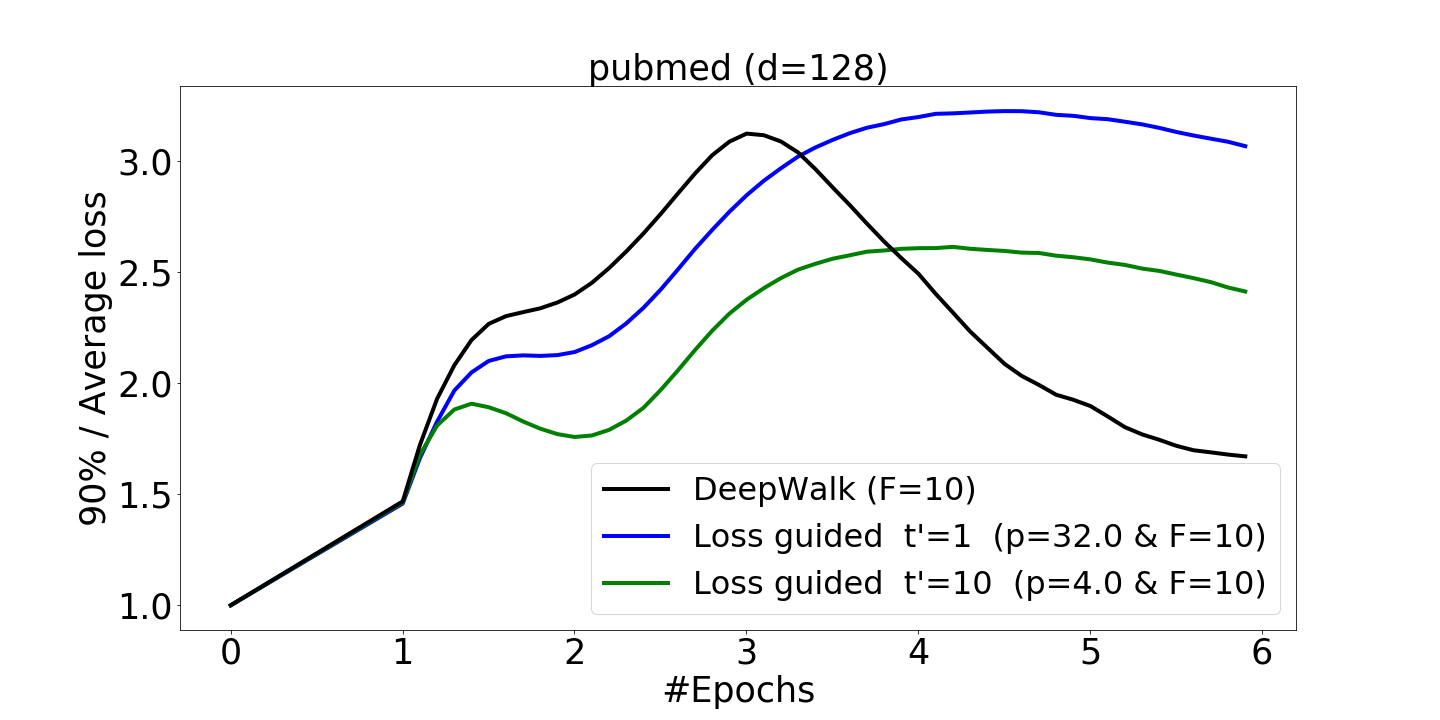}
 \includegraphics[width=0.65\columnwidth]{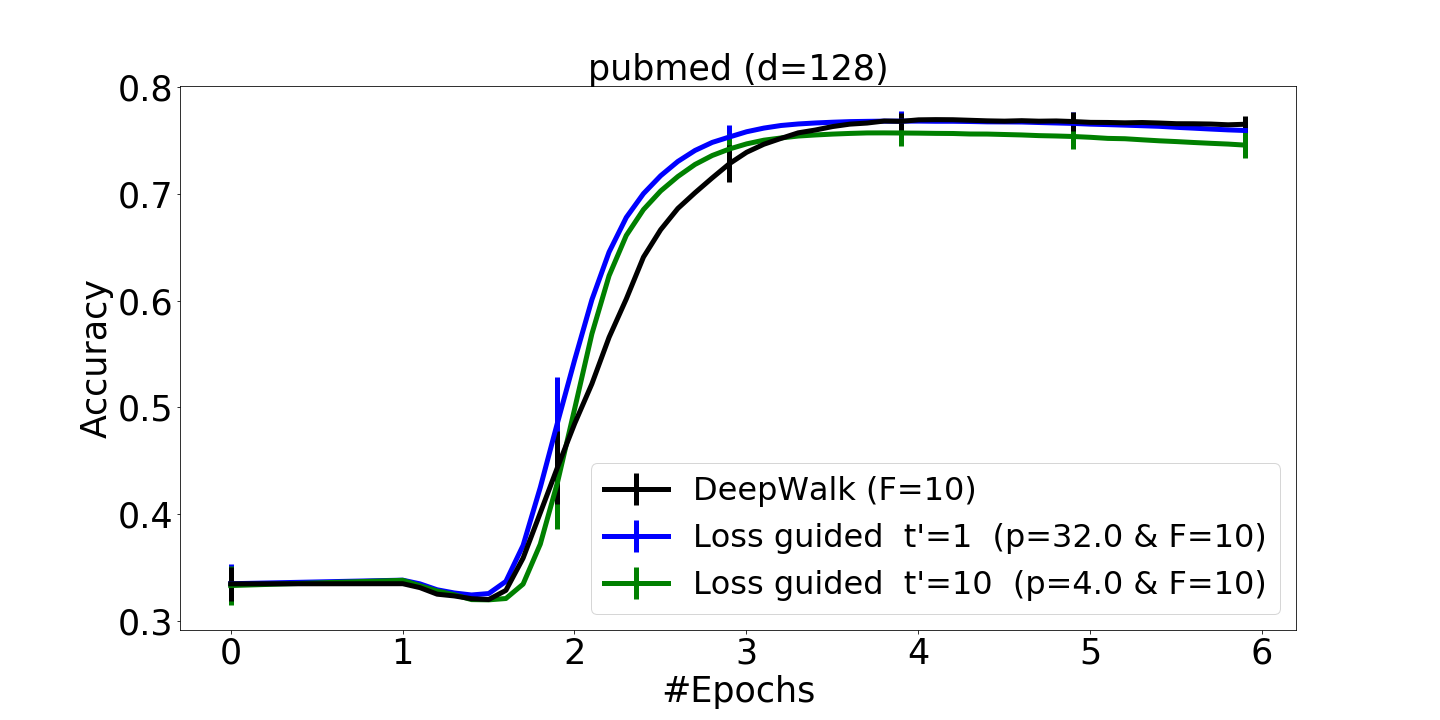}\\
 
 \includegraphics[width=0.65\columnwidth]{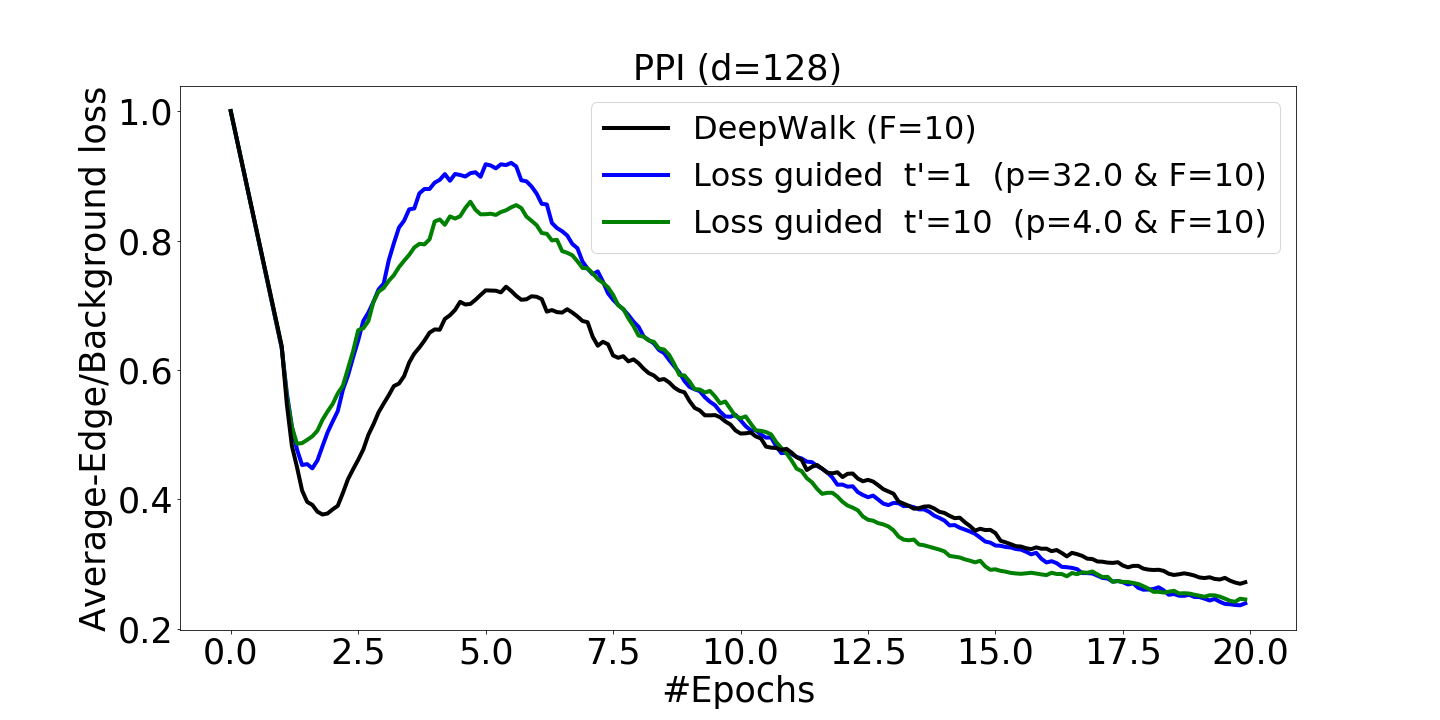}
 \includegraphics[width=0.65\columnwidth]{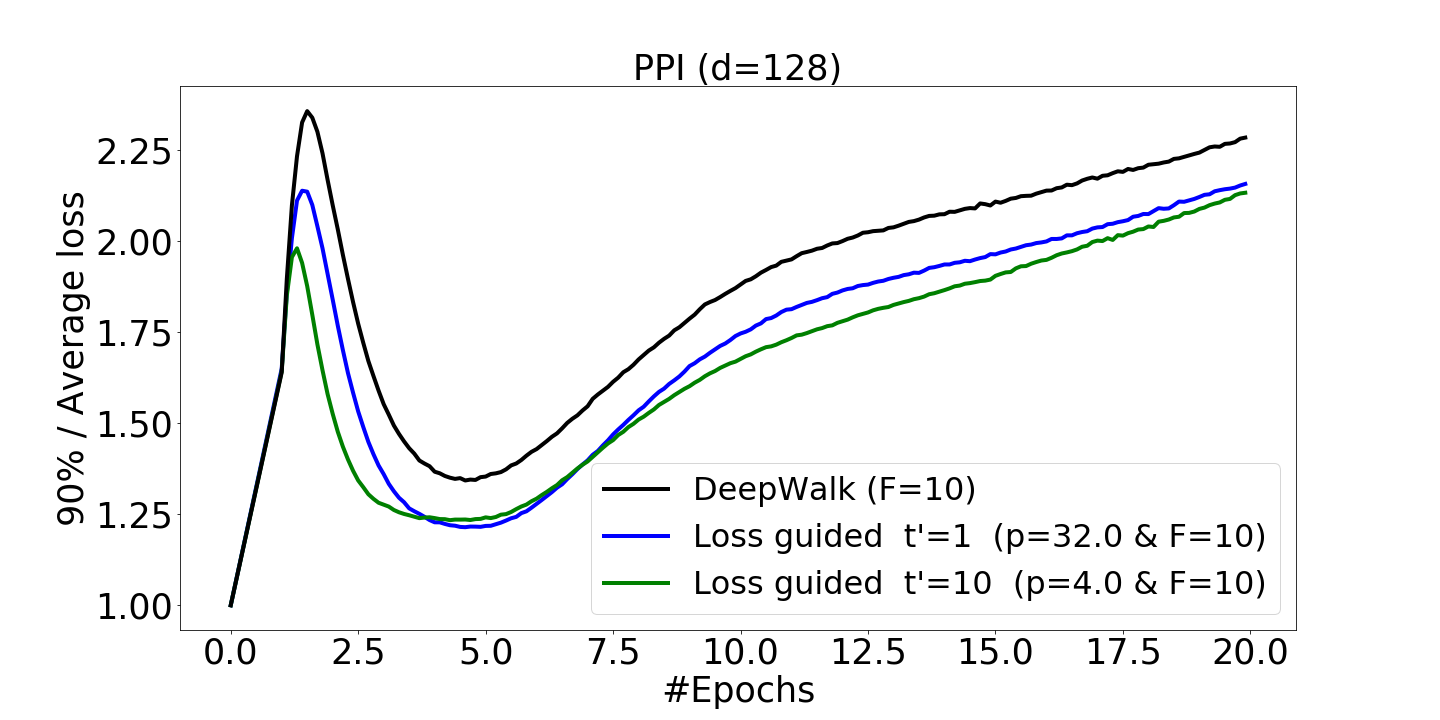}
 \includegraphics[width=0.65\columnwidth]{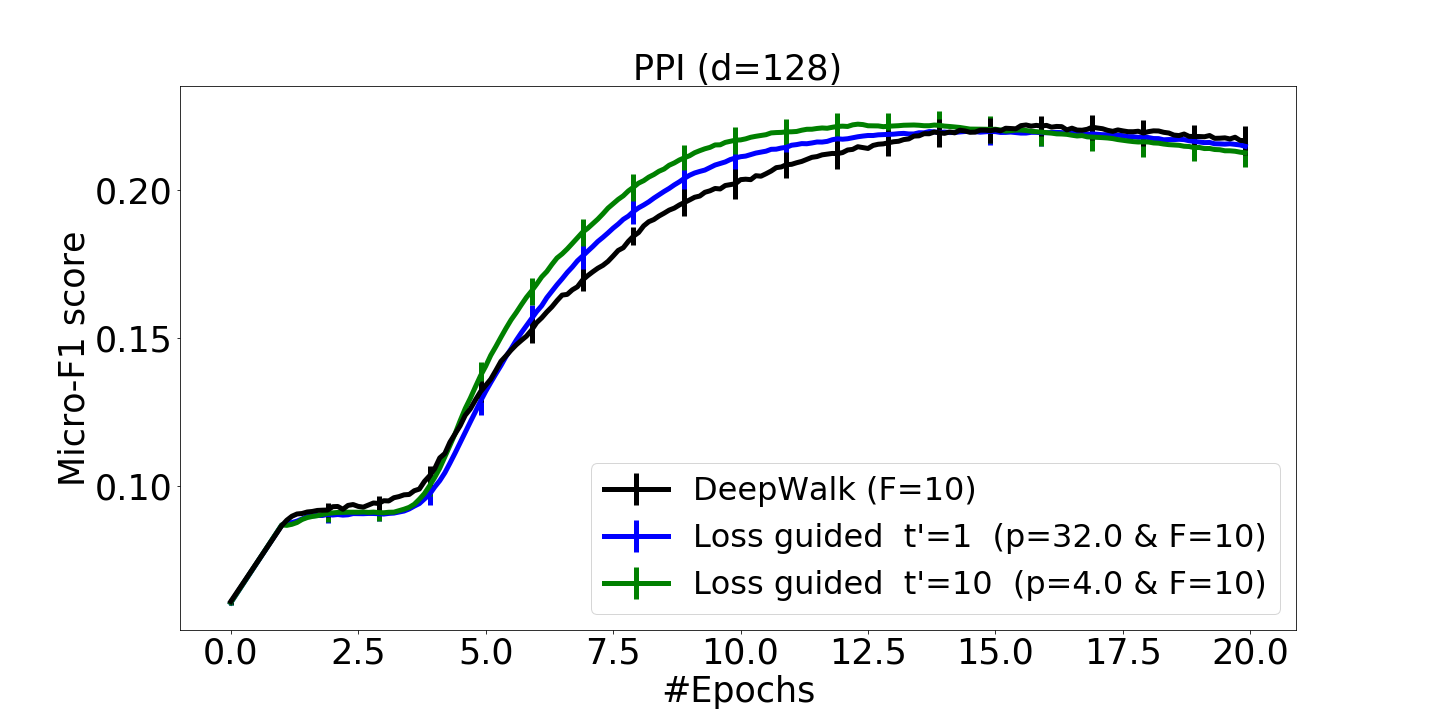}

\caption{Patterns of per-example loss values:  The ratio of edges loss versus background loss (left),  the 90\% quantile versus average edge loss (middle) and quality in the course of training (right).  For loss-guided methods with loss scoring $\Lscore_1$ and $\Lscore_{10}$ and the baseline method {\sc DeepWalk}. 
}
\label{plot:loss_and_perf}
\end{center}
\end{figure*}
\section{Loss behavior} \label{loss:sec}
We observed that loss-guided walk selection improves performance of the different downstream tasks.  To obtain insights on the behavior of the loss-guided versus the baseline methods we consider properties of the per-example loss values.  
Figure~\ref{plot:loss_and_perf} provides side-by-side plots of these properties and plots of quality in the course of training.
For our purposes here, we treat the graph edges as an approximate set of strong positive examples.  These examples tend to be weighted higher (have larger $\kappa_{i,j}$ values) in the distribution generated from random walks.  We consider two qualities of the distribution of the loss values $L_+(i,j)$ on these edges: 
\begin{trivlist}
\item[$\bullet$]
The ratio of the {\em average edge-loss}  to the {\em background} loss. The average edge-loss is the average $L_+(i,j)$ over graph edges and the background loss is measured by the average loss $L_+(i,j)$ over $10^3$ random non-edge pairs $(i,j)$.  We observed that the loss scale shifts significantly during training and in particular both these average loss values decrease by orders of magnitude.  The ratio serves as a normalized measures of separation between edge and background loss and we expect it to be lower (more separation) when the training is more effective.  
 \item[$\bullet$]
 The ratio of the 90\% quantile of edge loss values $L_+(i,j)$ to the average edge loss.   The ratio is a measure of spread and indicates how well the training method balances its  progress across the positive examples.  A ratio that is closer to 1 means a smaller spread and a better balance.
\end{trivlist}
Results on representative datasets are reported in Figure~\ref{plot:loss_and_perf}.
We can see that across datasets and in the training regime before performance peaks, the loss-guided methods
have a lower spread than the baseline {\sc DeepWalk} method:  The ratio of the 90\% percentile to the average loss on edges is uniformly lower.  Moreover, the loss-guided method with $\Lscore_{10}$ has a lower spread than $\Lscore_1$.  Overall, this is consistent with what we expect with loss-guided training, where more training is directed to higher loss examples and $\Lscore_{10}$ better representing the current loss than $\Lscore_1$.

Interestingly,  the baseline method {\sc DeepWalk} has a lower ratio, which corresponds to stronger separation of edge-loss and background loss.  The lower ratio of the baseline starts early on and surprisingly perhaps, on some of the datasets (PPI and the citation networks), persists 
also in regimes where {\sc DeepWalk} is outperformed by the loss-guided  methods.

These patterns showcase the advantage of loss-guided selections that are more geared to minimize spread rather than average loss. The average loss seems to indeed be effectively minimized by baseline methods, but on its own does not fully reflects on quality.  

\end{document}